\documentclass[11pt]{article}

\usepackage[final]{acl}

\usepackage{times}
\usepackage{latexsym}

\usepackage[T1]{fontenc}

\usepackage[utf8]{inputenc}

\usepackage{microtype}

\usepackage{inconsolata}

\usepackage{graphicx}

\usepackage{booktabs}
\usepackage{colortbl}
\usepackage{xcolor}
\usepackage{amssymb}
\usepackage{amsmath}
\usepackage{tcolorbox}
\usepackage{algorithm}
\usepackage{algpseudocode}
\usepackage{textcomp}
\usepackage{hyperref}

\definecolor{darkgreen}{HTML}{006400}

\definecolor{lightgreen}{HTML}{C2F2CE}
\definecolor{lightred}{HTML}{FFCCC9}
\definecolor{lightblue}{HTML}{C2F0FC}
\definecolor{Darkgreen}{HTML}{2CCC23}
\definecolor{Darkblue}{HTML}{26A1FF}
\definecolor{Gray}{HTML}{EDEDED}

\newcommand{\mycolorbox}[2]{\begingroup\fboxsep=1pt\colorbox{#1}{#2}\endgroup}

\usepackage{scalerel}
\def\hf{\scalerel*{\includegraphics{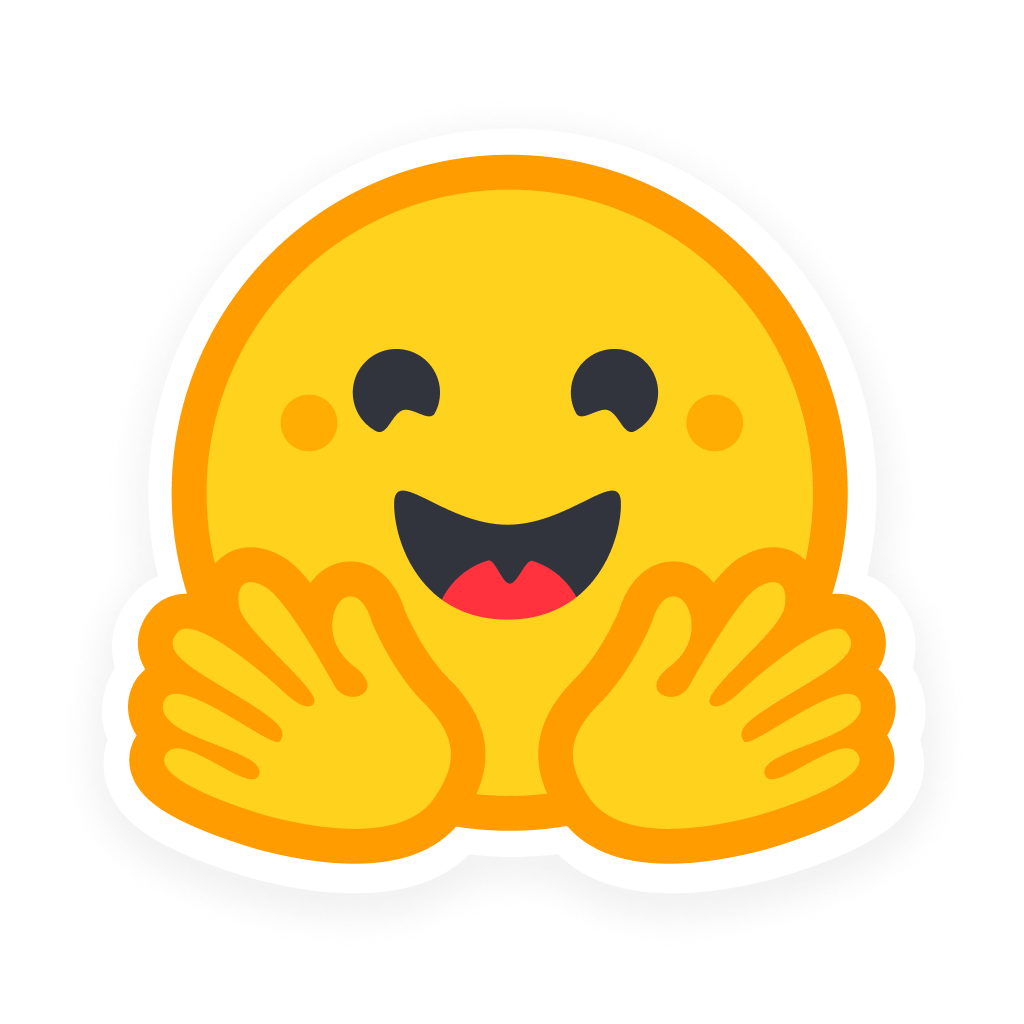}}{\textrm{\textbigcircle}}}
\def\github{\scalerel*{\includegraphics{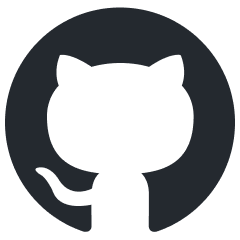}}{\textrm{\textbigcircle}}}

\title{SeLaR: Selective Latent Reasoning in Large Language Models}

\makeatletter
\def\thanks#1{\protected@xdef\@thanks{\@thanks
        \protect\footnotetext{#1}}}
\makeatother
\author{
Renyu Fu \qquad Guibo Luo$^{\dag}$\thanks{† Corresponding author.} \\
{\normalsize Guangdong Provincial Key Laboratory of Ultra High Definition Immersive Media Technology} \\
{\normalsize Shenzhen Graduate School, Peking University} \\
\href{mailto:luogb@pku.edu.cn}{\texttt{luogb@pku.edu.cn}} \\
\github\ \href{https://github.com/Parker-rfu/SeLaReasoning}{github.com/Parker-rfu/SeLaReasoning}
}

\begin{document}
\maketitle
\begin{abstract}
Chain-of-Thought (CoT) has become a cornerstone of reasoning in large language models, yet its effectiveness is constrained by the limited expressiveness of discrete token sampling. Recent latent reasoning approaches attempt to alleviate this limitation by replacing discrete tokens with soft embeddings (probability-weighted mixtures of token embeddings) or hidden states, but they commonly suffer from two issues: (1) global activation injects perturbations into high-confidence steps, impairing reasoning stability; and (2) soft embeddings quickly collapse toward the highest-probability token, limiting exploration of alternative trajectories. To address these challenges, we propose \textit{\textbf{SeLaR}} (\textbf{Se}lective \textbf{La}tent \textbf{R}easoning), a lightweight and training-free framework. SeLaR introduces an entropy-gated mechanism that activates soft embeddings only at low-confidence steps, while preserving discrete decoding at high-confidence steps. Additionally, we propose an entropy-aware contrastive regularization that pushes soft embeddings away from the highest-probability token's direction, encouraging sustained exploration of multiple latent reasoning paths. Experiments on five reasoning benchmarks demonstrate that SeLaR consistently outperforms standard CoT and state-of-the-art training-free methods.
\end{abstract}

\section{Introduction}
Chain-of-Thought (CoT) \cite{wei2023chainofthoughtpromptingelicitsreasoning,goyal2024thinkspeaktraininglanguage,pfau2024letsthinkdotdot} has become a prevailing paradigm for enabling multi-step reasoning in large language models \cite{brown2020languagemodelsfewshotlearners,chowdhery2022palmscalinglanguagemodeling,du2022glmgenerallanguagemodel,touvron2023llamaopenefficientfoundation,openai2024gpt4technicalreport,singh2025openaigpt5card}. By explicitly generating intermediate reasoning steps, CoT significantly improves performance on complex tasks such as mathematical and logical reasoning \cite{deepseekai2025deepseekr1incentivizingreasoningcapability,openai2024openaio1card,openai2025gptoss120bgptoss20bmodel,abdin2025phi4reasoningtechnicalreport,qwen2025qwen25technicalreport,kimiteam2025kimik15scalingreinforcement}. However, CoT relies on hard token commitments at each step: the model must discretize its internal distribution into a single sampled token, potentially discarding valuable information about alternative reasoning paths. This commitment may hinder the effective propagation of uncertainty across reasoning steps, ultimately leading to suboptimal final predictions \cite{li2025implicitreasoninglargelanguage}.

\begin{figure*}[t]
  \centering
  \includegraphics[width=\linewidth]{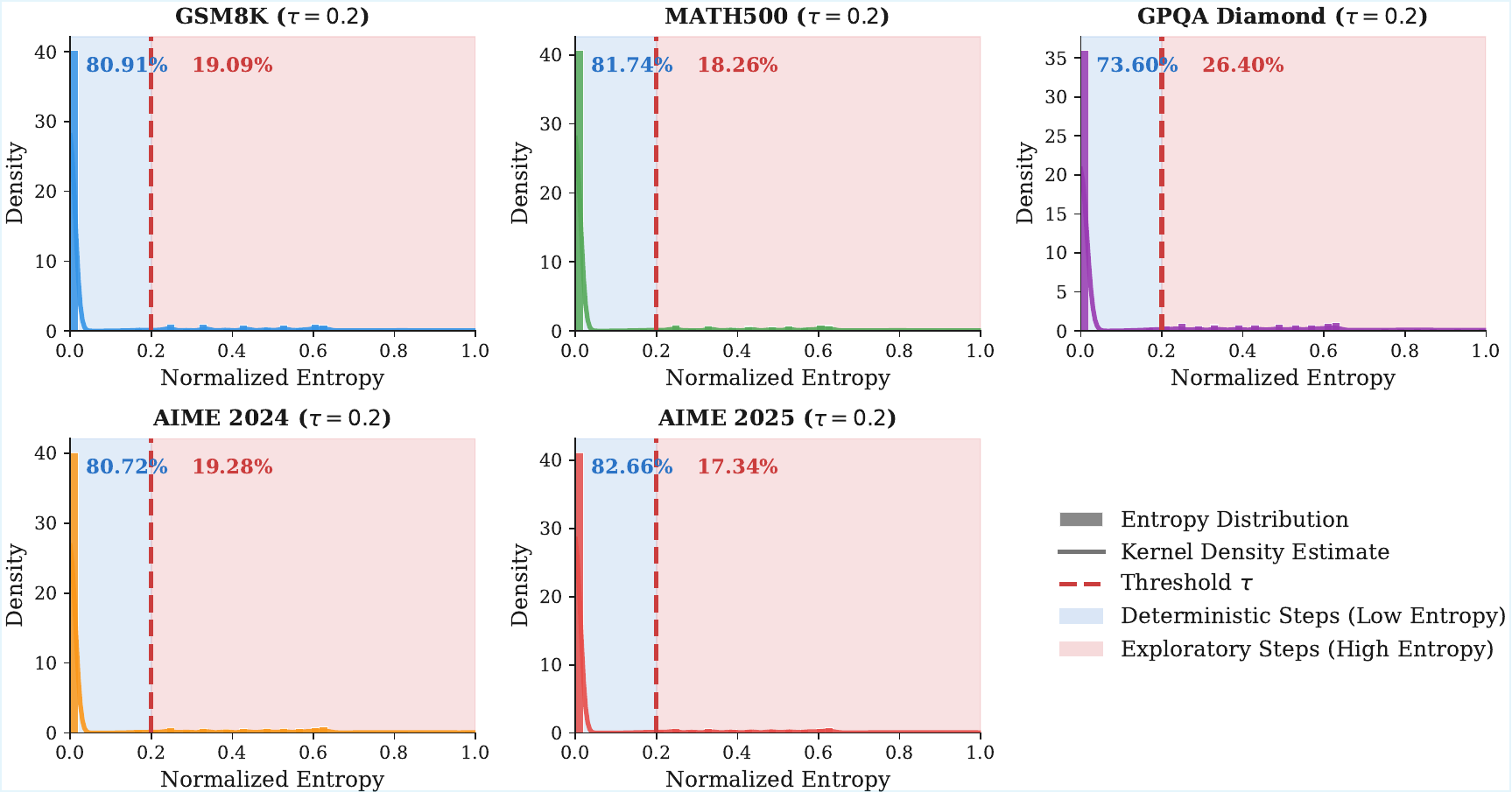}
  \caption{Normalized entropy distributions of decoding steps across five reasoning benchmarks using Qwen3-8B. Each subplot shows the density of step-wise entropy values, revealing a clear long-tail structure: the majority of steps concentrate in a low-entropy region (deterministic steps), while a smaller tail extends toward higher entropy values (exploratory steps).}
  \label{fig:entropy-distribution}
\end{figure*}

Inspired by human reasoning, which often considers multiple plausible hypotheses simultaneously, recent work has explored latent reasoning paradigms that replace discrete token sampling with soft embeddings or hidden states as carriers of reasoning information \cite{hao2025traininglargelanguagemodels,cheng2024compressedchainthoughtefficient,xu2025softcotsoftchainofthoughtefficient,zhang2025softthinkingunlockingreasoning,tan2025thinksilentlythinkfast,shi2025swireasoningswitchthinkinglatentexplicit}. These approaches enable richer representations and implicit branching over multiple candidate tokens during reasoning. 

Existing latent reasoning methods can be categorized into fine-tuning–based and training-free approaches. Fine-tuning methods such as Coconut~\cite{hao2025traininglargelanguagemodels} propagate hidden states as reasoning signals, but often suffer from catastrophic forgetting~\cite{lobo2025impactfinetuningchainofthoughtreasoning} due to the domain gap between hidden-state and input embedding spaces. Training-free methods such as Soft Thinking~\cite{zhang2025softthinkingunlockingreasoning} employ soft embeddings to explore multiple reasoning trajectories, but activate them uniformly across all steps regardless of model confidence.

Our work is motivated by a key empirical observation: during CoT decoding, the entropy of the model’s output distribution exhibits a clear \textbf{\textit{long-tail}} pattern across reasoning steps. As illustrated in Figure~\ref{fig:entropy-distribution}, most reasoning steps cluster in a low-entropy region, reflecting confident token commitments, while a small but consistent tail extends to higher entropy values, corresponding to moments of increased ambiguity. We refer to the former as \textit{\textbf{deterministic steps}}, where the model decisively commits to a single token, and the latter as \textit{\textbf{exploratory steps}}, where multiple candidates compete and broader exploration may be beneficial.

This entropy-based view reveals a key limitation of existing latent reasoning methods: global activation ignores the long-tail structure of model confidence, applying soft embeddings indiscriminately to both deterministic and exploratory steps. At deterministic steps where the model is already confident, this introduces unnecessary perturbations that undermine reasoning stability. Furthermore, even at exploratory steps, prior work~\cite{wu2025llmssinglethreadedreasonersdemystifying} shows that soft embeddings collapse prematurely toward the dominant token, concentrating reasoning on a single trajectory and suppressing alternatives.

To address these limitations, we propose \textit{\textbf{SeLaR}}, a selective and training-free latent reasoning framework. This paper centers on two key questions: \textit{\textbf{(i) When should latent reasoning be activated?}} SeLaR introduces an entropy-gated mechanism that activates soft embeddings only at exploratory steps, while preserving discrete decoding at deterministic steps. \textit{\textbf{(ii) How can premature collapse toward the dominant token be mitigated?}} SeLaR incorporates an entropy-aware contrastive regularization that pushes soft embeddings away from the dominant token's direction in proportion to entropy magnitude, sustaining exploration across alternative reasoning paths. Our contributions are summarized as follows:
\begin{itemize}
    \item We empirically show that only a small fraction of reasoning steps exhibit high uncertainty, and that activating latent reasoning exclusively at exploratory steps significantly outperforms globally applied latent reasoning.
    \item We propose SeLaR, a lightweight and training-free framework that selectively activates latent reasoning via entropy gating at exploratory steps, while preserving standard discrete decoding at deterministic steps. To further prevent premature collapse toward the dominant token, SeLaR introduces an entropy-aware contrastive regularization that sustains multiple latent reasoning alternatives.
    \item Extensive experiments on five reasoning benchmarks across multiple model scales demonstrate that SeLaR consistently outperforms standard CoT decoding and state-of-the-art training-free reasoning methods.
\end{itemize}
\section{Related Work}
\subsection*{Chain-of-Thought Reasoning}
Chain-of-Thought (CoT) reasoning enhances the problem-solving capabilities of large language models by explicitly generating intermediate reasoning steps, and has become a central paradigm for improving multi-step reasoning \cite{zhou2023leasttomostpromptingenablescomplex,shinn2023reflexionlanguageagentsverbal,madaan2023selfrefineiterativerefinementselffeedback,zheng2024progressivehintpromptingimprovesreasoning,wang2024mathshepherdverifyreinforcellms,havrilla2024teachinglargelanguagemodels,shao2024deepseekmathpushinglimitsmathematical,chu2024navigateenigmaticlabyrinthsurvey,wang2024guidinglanguagemodelreasoning,saunshi2024inductivebiasstackingimproving,jin2025disentanglingmemoryreasoningability,wei2025surveyfeedbackbasedmultistepreasoning,yu2025flowreasoningtrainingllms,lee2025evolvingdeeperllmthinking}. Subsequent studies have primarily focused on improving CoT through decoding and search strategies. For example, self-consistency \cite{wang2023selfconsistencyimproveschainthought} mitigates the instability of single reasoning paths by sampling multiple trajectories and aggregating their predictions, while tree- \cite{yao2023reactsynergizingreasoningacting} or graph-based \cite{Besta_2024} search methods explicitly explore multiple discrete reasoning paths to improve robustness. Despite its strong empirical performance, CoT operates by committing to a single sequence of discrete tokens at each step, which can obscure or eliminate information about other plausible reasoning trajectories.
\subsection*{Latent Reasoning}
Latent reasoning differs from explicit CoT by leveraging hidden states or soft embeddings \cite{deng2023implicitchainthoughtreasoning,geiping2025scalingtesttimecomputelatent,yang2025largelanguagemodelslatently,shalev2024distributionalreasoningllmsparallel,mohtashami2024cotformerchainofthoughtdrivenarchitecture,wu2026parallelcontinuouschainofthoughtjacobi,wang2025system15reasoningtraversallanguage,su2025tokenassortedmixinglatent,zhang2025lightthinkerthinkingstepbystepcompression} to convey intermediate reasoning signals. Prior work in this area generally falls into two categories. Fine-tuning-based methods \cite{hao2025traininglargelanguagemodels,cheng2024compressedchainthoughtefficient,xu2025softcotsoftchainofthoughtefficient} propagate hidden states across reasoning steps via full or partial fine-tuning, enabling implicit multi-step reasoning that goes beyond discrete token generation.
In contrast, training-free methods \cite{zhang2025softthinkingunlockingreasoning,wu2025llmssinglethreadedreasonersdemystifying,shi2025swireasoningswitchthinkinglatentexplicit} replace discrete token inputs with probability-weighted soft embeddings, allowing models to operate in continuous space without parameter updates. Our approach belongs to the latter category, but diverges from existing paradigms that apply latent reasoning globally. Specifically, SeLaR employs an entropy-gated mechanism to selectively activate latent reasoning only at exploratory steps, while incorporating a contrastive regularization strategy to prevent premature collapse toward the dominant token's trajectory during the reasoning process.
\section{Method}
\subsection{Overview}
We propose SeLaR, a selective and training-free latent reasoning framework that dynamically activates latent reasoning only when necessary. The core idea is to avoid globally propagating soft embeddings throughout the entire decoding process. Instead, SeLaR leverages token-level entropy as a confidence signal to identify high-uncertainty exploratory steps, at which latent reasoning is selectively enabled. For deterministic steps where the model exhibits high confidence, standard discrete decoding is preserved to maintain stability and efficiency.

\begin{figure*}[t]
  \centering
  \includegraphics[width=\linewidth]{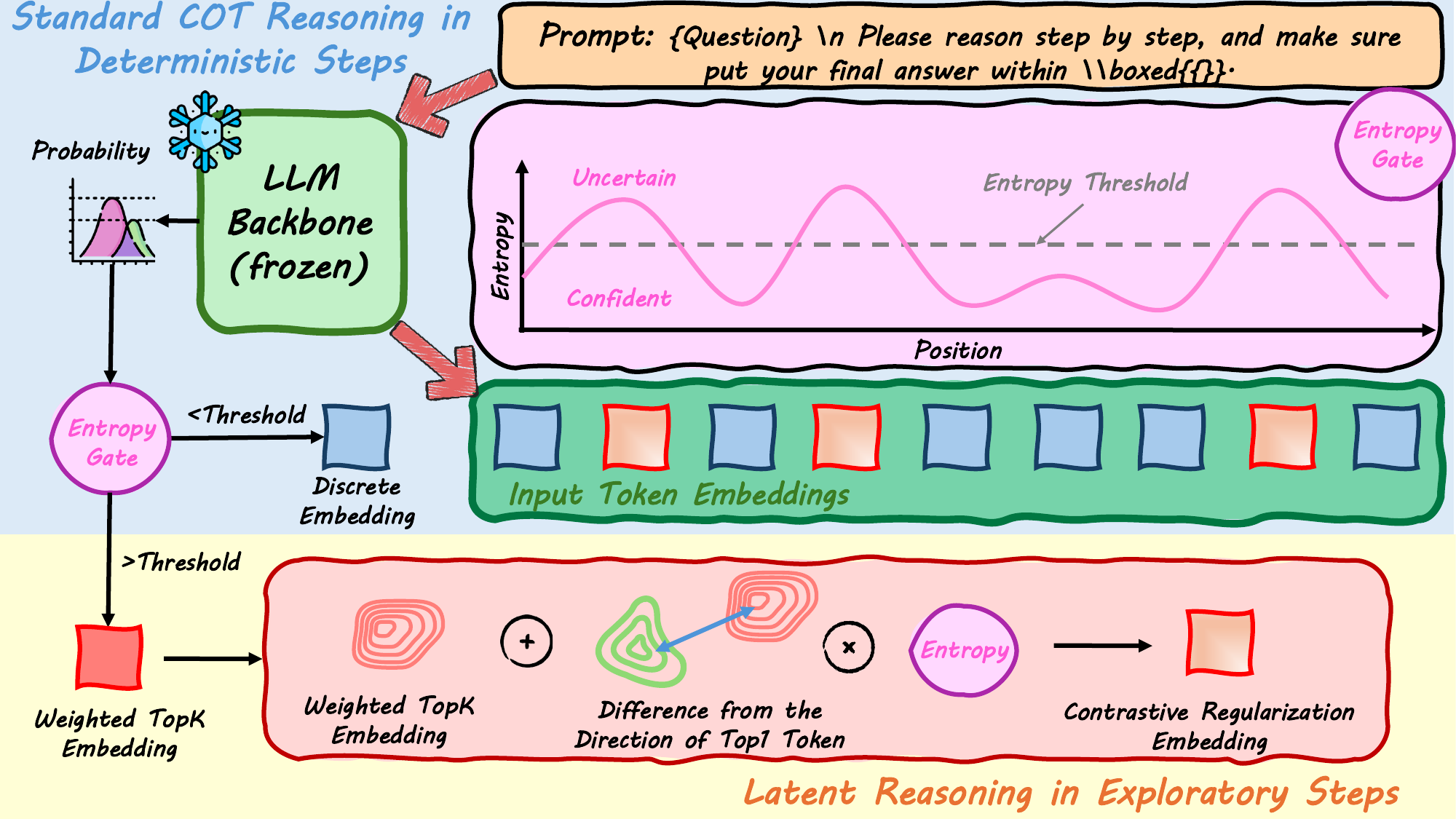}
  \caption{
    Overview of SeLaR. At each decoding step, we compute the normalized entropy over top-$k$ tokens. If entropy falls below threshold $\tau$ (deterministic step), standard discrete decoding is applied. Otherwise (exploratory step), we construct a soft embedding from top-$k$ candidates and apply contrastive regularization to push it away from the dominant token, encouraging exploration of alternative reasoning paths.
  }
  \label{fig:SeLaR_Framework}
\end{figure*}

As shown in Figure~\ref{fig:SeLaR_Framework}, SeLaR comprises two components: (1) an \textbf{\textit{entropy-gated selective mechanism}} that determines when latent reasoning should be activated during decoding, and (2) an \textbf{\textit{entropy-aware contrastive regularization}} that mitigates the tendency of soft embeddings to overemphasize the highest-probability token, which increasingly dominates subsequent predictions and suppresses alternative reasoning paths.

\subsection{Background}
\paragraph{Standard Chain-of-Thought Reasoning}
Given an input query $q$, a language model $\mathcal{L}$ generates a reasoning sequence $r_{1:T} = (x_1, \ldots, x_T)$ followed by a final answer $a$. At each decoding step $t$, the model produces a conditional distribution over the vocabulary $\mathcal{V}$:
\begin{equation}
p_t(v) = p(v \mid q, x_{<t}), \quad v \in \mathcal{V}.
\end{equation}
Standard Chain-of-Thought (CoT) decoding commits to a single discrete token $x_t$ at each step:
\begin{equation}
x_t =
\begin{cases}
\displaystyle \arg\max_{v \in \mathcal{V}} p_t(v), 
& \textit{Greedy} \\[6pt]
\displaystyle v \sim \tilde{p}_t(v), 
& \textit{Sampling}
\end{cases}
\end{equation}
where $\tilde{p}_t$ is the filtered distribution obtained by applying temperature scaling and truncation strategies (e.g., top-$k$, top-$p$) to the original distribution $p_t$. The embedding $e_{x_t}$ is then used as input for the next decoding step.

\paragraph{Latent Reasoning with Soft Embeddings}

Latent reasoning methods replace discrete token inputs with soft embeddings to preserve distributional information. Let $E \in \mathbb{R}^{|\mathcal{V}| \times d}$ denote the embedding matrix. At step $t$, instead of committing to a sampled token, a soft embedding is computed as:
\begin{equation}
\label{eq:soft-embedding}
e_t = \sum_{v \in \mathcal{V}} p_t(v) \cdot e_v = \sum_{v \in \mathcal{V}} p_t(v) \cdot E_v.
\end{equation}
This soft embedding is fed to the model as the next-step input, enabling implicit exploration of multiple candidate tokens within a single forward pass.

\subsection{Entropy-gated Selective Mechanism}

\paragraph{Motivation.}
Existing training-free latent reasoning methods propagate soft embeddings at every decoding step. While enabling richer representations, this global activation injects unnecessary perturbation into confident steps, undermining reasoning stability. Our key insight is that \textbf{\textit{latent reasoning is only necessary when the model is uncertain}}, motivating a selective mechanism that activates it only at critical exploratory steps.

\paragraph{Entropy as a Measure of Uncertainty}
At decoding step $t$, the model produces a predictive distribution $p_t(\cdot)$ over the vocabulary $\mathcal{V}$. Rather than computing entropy over the full vocabulary as in prior work \cite{shi2025swireasoningswitchthinkinglatentexplicit}, we estimate uncertainty using the top-$k$ most probable tokens, which dominate the model’s predictive mass and are most relevant for decision making. Specifically, let $\mathcal{V}_k \subset \mathcal{V}$ denote the set of top-$k$ tokens under $p_t$. We first renormalize the distribution over $\mathcal{V}_k$:
\begin{equation}
\hat{p}_t(v) = \frac{p_t(v)}{\sum_{u \in \mathcal{V}_k} p_t(u)}, \quad v \in \mathcal{V}_k,
\end{equation}
and define the truncated entropy as:
\begin{equation}
H_t = - \sum_{v \in \mathcal{V}_k} \hat{p}_t(v) \log \hat{p}_t(v),
\end{equation}

\begin{equation}
\bar{H}_t = \mathrm{clamp}\left( \frac{H_t}{\log k},\; 0,\; 1 \right).
\end{equation}
This top-$k$ entropy captures the model’s uncertainty among its most plausible candidates while avoiding perturbation from the low-probability tokens. Low entropy indicates confident predictions dominated by a small number of candidates, whereas high entropy reflects ambiguity among multiple competing tokens.

\paragraph{Threshold Selection}
The entropy threshold $\tau$ determines when latent reasoning is activated. Across models and benchmarks, $\bar{H}_t$ exhibits a clear long-tail pattern during CoT decoding: a dominant low-entropy region where a single token commands the predictive mass (deterministic steps), and a sparse long-tail high-entropy region where multiple tokens compete (exploratory steps). The low-density transition between these two regions marks a qualitative shift from single-token dominance to multi-token competition, providing a principled and natural boundary for selecting $\tau$. Consequently, $\tau$ is positioned within this transition band, serving as a separator that demarcates high-confidence deterministic steps from low-confidence exploratory ones. As shown in Appendix~\ref{app:Sensitivity Analysis Details}, SeLaR is robust to the exact choice of $\tau$, with stable performance across $\tau \in [0.3, 0.7]$.

\paragraph{Entropy-Gated Selective Activation}
Given the entropy threshold $\tau$, the input for the next step is then computed as:
\begin{equation}
e_t =
\begin{cases}
E_{x_t}, & \text{if } \bar{H}_t \leq \tau, \\[4pt]
\displaystyle\sum_{v \in \mathcal{V}_k} \hat{p}_t(v) \cdot e_v, & \text{if } \bar{H}_t > \tau,
\end{cases}
\end{equation}
where $e_v$ denotes the embedding of token $v$. At deterministic steps, the model follows standard discrete decoding by committing to a single sampled token. At exploratory steps, latent reasoning is activated by replacing the discrete token with a soft embedding. This entropy-gated mechanism enables latent reasoning only when it is most beneficial, while maintaining the stability of standard decoding elsewhere.

\subsection{Entropy-aware Contrastive Regularization}

\paragraph{Motivation}
Selective activation addresses \emph{\textbf{when}} to apply latent reasoning, but does not address \emph{\textbf{how}} to maintain effective exploration once activated. We now turn to a complementary challenge: preventing soft embeddings from prematurely collapsing back to a single token during the reasoning process.

\paragraph{Premature Collapse in Latent Reasoning}
Although soft embeddings enable implicit exploration of multiple candidate tokens, prior work~\cite{wu2025llmssinglethreadedreasonersdemystifying} has identified a \emph{premature collapse} phenomenon: during latent reasoning, soft embeddings quickly become dominated by the highest-probability token, effectively degenerating to greedy decoding. Formally, let $v_t^{\ast} = \arg\max_{v \in \mathcal{V}_k} \hat{p}_t(v)$ denote the dominant token at step $t$. The soft embedding $e_t$ tends to align increasingly with $e_{v_t^{\ast}}$ as decoding proceeds. This alignment accelerates convergence toward a single trajectory, undermining the multi-path exploration that soft embeddings are designed to enable.

\paragraph{Entropy-aware Contrastive Regularization}
To counteract premature collapse, we introduce a contrastive regularization that explicitly pushes the soft embedding away from the dominant token direction. At each exploratory step, we compute the difference between the soft embedding and the dominant token embedding:
\begin{equation}
\Delta_t = e_t - e_{v_t^{\ast}}, \quad \hat{\Delta}_t = \frac{\Delta_t}{\|\Delta_t\| + \epsilon},
\end{equation}
where $\hat{\Delta}_t$ is the unit direction pointing from $e_{v_t^{\ast}}$ toward $e_t$. The regularized soft embedding is then computed as:
\begin{equation}
\tilde{e}_t = e_t + \bar{H}_t \cdot \hat{\Delta}_t \cdot \|\Delta_t\|.
\end{equation}
This formulation scales the repulsion from the dominant token according to the model’s uncertainty: when entropy is high, the regularization effect is strong, encouraging broader exploration; as the model becomes confident, the effect diminishes naturally.

\section{Experiments}
\subsection{Settings}
\paragraph{Datasets}
We conduct experiments on five reasoning datasets, including GSM8K \cite{cheng2024compressedchainthoughtefficient}, MATH500 \cite{hendrycks2021measuringmathematicalproblemsolving}, AIME2024 \cite{huggingfaceh42024aime}, AIME2025 \cite{yentinglin2025aime} in the mathematical domain, and GPQA-Diamond \cite{rein2024gpqa} for STEM reasoning. For more details, please refer to Appendix \ref{sec:datasets}.
\paragraph{Baselines}
We compare SeLaR against four baselines: (1) standard CoT reasoning with sampling, (2) standard CoT reasoning with greedy decoding, (3) Soft Thinking~\cite{zhang2025softthinkingunlockingreasoning}, a training-free latent reasoning method that globally applies soft embeddings, and (4) SwiReasoning (SwiR)~\cite{shi2025swireasoningswitchthinkinglatentexplicit}, which switches between explicit and latent reasoning modes based on the relative entropy increase between adjacent decoding steps. However, relying on between-step entropy deltas makes the trigger prone to spurious firing, forcing SwiR to resort to window-based smoothing heuristics that introduce additional hyperparameters. 
\begin{table*}[ht]
\centering
\caption{
Detailed results on reasoning benchmarks.
Results highlighted in \mycolorbox{lightgreen}{green} indicate performance comparable to or better than CoT (Sampling).
Results highlighted in \mycolorbox{lightred}{red} indicate a performance drop relative to CoT (Sampling).
}
\label{tab:bridge_results}

\begin{tabular}{lcccccc}
\toprule
\multicolumn{1}{l|}{Method} &
GSM8K &
MATH500 &
GPQA &
AIME24 &
AIME25 &
Avg \\ \midrule

\multicolumn{1}{l|}{} &
\multicolumn{6}{c}{\cellcolor{lightblue}\textit{Qwen3-1.7B} \cite{yang2025qwen3technicalreport}} \\
\multicolumn{1}{l|}{CoT (Sampling)} & 
90.07 & 92.00 & \textbf{39.39} & 50.00 & 33.33 & {\color[HTML]{333333}60.96} \\
\multicolumn{1}{l|}{CoT (Greedy)} & 
\cellcolor{lightred}88.32 &
\cellcolor{lightred}90.60 &
\cellcolor{lightred}31.31 &
\cellcolor{lightred}40.00 &
\cellcolor{lightred}30.00 &
\cellcolor{lightred}56.05 \\
\multicolumn{1}{l|}{Soft Thinking} &
\cellcolor{lightred}89.46 &
\cellcolor{lightred}91.00 &
\cellcolor{lightred}33.83 &
\cellcolor{lightred}36.67 &
\cellcolor{lightgreen}\textbf{36.67} &
\cellcolor{lightred}57.53 \\
\multicolumn{1}{l|}{SwiR} &
\cellcolor{lightgreen}89.84 &
\cellcolor{lightred}92.00 &
\cellcolor{lightred}37.88 &
\cellcolor{lightred}46.67 &
\cellcolor{lightred}23.33 &
\cellcolor{lightred}57.94 \\
\multicolumn{1}{l|}{\textbf{SeLaR}} &
\cellcolor{lightgreen}\textbf{90.60} &
\cellcolor{lightgreen}\textbf{92.60} &
\cellcolor{lightred}35.35 &
\cellcolor{lightgreen}\textbf{53.33} &
\cellcolor{lightgreen}\textbf{36.67} &
\cellcolor{lightgreen}\textbf{61.71} \\ \midrule

\multicolumn{1}{l|}{} &
\multicolumn{6}{c}{\cellcolor{lightblue}\textit{Qwen3-8B} \cite{yang2025qwen3technicalreport}} \\
\multicolumn{1}{l|}{CoT (Sampling)} &
95.45 & \textbf{98.00} & 61.62 & 76.67 & 66.67 & {\color[HTML]{333333}79.68} \\
\multicolumn{1}{l|}{CoT (Greedy)} &
\cellcolor{lightred}95.22 &
\cellcolor{lightred}96.20 &
\cellcolor{lightred}55.05 &
\cellcolor{lightred}70.00 &
\cellcolor{lightred}63.33 &
\cellcolor{lightred}75.96 \\
\multicolumn{1}{l|}{Soft Thinking} &
\cellcolor{lightred}94.92 &
\cellcolor{lightred}95.80 &
\cellcolor{lightred}57.58 &
\cellcolor{lightred}70.00 &
\cellcolor{lightgreen}66.67 &
\cellcolor{lightred}76.99 \\
\multicolumn{1}{l|}{SwiR} &
\cellcolor{lightgreen}95.68 &
\cellcolor{lightred}97.00 &
\cellcolor{lightgreen}\textbf{62.63} &
\cellcolor{lightred}60.00 &
\cellcolor{lightgreen}66.67 &
\cellcolor{lightred}76.40 \\
\multicolumn{1}{l|}{\textbf{SeLaR}} &
\cellcolor{lightgreen}\textbf{95.83} &
\cellcolor{lightred}97.00 &
\cellcolor{lightgreen}{61.62} &
\cellcolor{lightgreen}\textbf{83.33} &
\cellcolor{lightgreen}\textbf{80.00} &
\cellcolor{lightgreen}\textbf{83.56} \\ \midrule

\multicolumn{1}{l|}{} &
\multicolumn{6}{c}{\cellcolor{lightblue}\textit{Qwen3-32B} \cite{yang2025qwen3technicalreport}} \\
\multicolumn{1}{l|}{CoT (Sampling)} &
95.83 & 97.40 & 66.16 & 80.42 & 72.08 & {\color[HTML]{333333}82.38} \\
\multicolumn{1}{l|}{CoT (Greedy)} &
\cellcolor{lightgreen}95.91 &
\cellcolor{lightred}97.20 &
\cellcolor{lightgreen}69.70 &
\cellcolor{lightred}80.00 &
\cellcolor{lightgreen}73.33 &
\cellcolor{lightgreen}83.23 \\
\multicolumn{1}{l|}{Soft Thinking} &
\cellcolor{lightred}95.75 &
\cellcolor{lightred}97.40 &
\cellcolor{lightgreen}67.17 &
\cellcolor{lightred}74.58 &
\cellcolor{lightred}66.25 &
\cellcolor{lightred}80.23 \\
\multicolumn{1}{l|}{SwiR} &
\cellcolor{lightgreen}\textbf{96.21} &
\cellcolor{lightgreen}\textbf{98.40} &
\cellcolor{lightgreen}\textbf{70.20} &
\cellcolor{lightgreen}82.92 &
\cellcolor{lightgreen}73.75 &
\cellcolor{lightgreen}84.30 \\
\multicolumn{1}{l|}{\textbf{SeLaR}} &
\cellcolor{lightgreen}96.06 &
\cellcolor{lightgreen}97.60 &
\cellcolor{lightgreen}67.17 &
\cellcolor{lightgreen}\textbf{83.33} &
\cellcolor{lightgreen}\textbf{80.00} &
\cellcolor{lightgreen}\textbf{84.83} \\
\bottomrule
\end{tabular}
\end{table*}
\paragraph{Implementation Details.}
We evaluate SeLaR on three reasoning-oriented LLMs: Qwen3-1.7B, Qwen3-8B, and Qwen3-32B~\cite{yang2025qwen3technicalreport}. All experiments are implemented using the Hugging Face Transformers framework \cite{wolf2020huggingfacestransformersstateoftheartnatural}. For fair comparison, all baselines are reproduced under identical hardware conditions (4$\times$ NVIDIA RTX PRO 6000 GPUs) using the official implementation and reported hyperparameters. Since the SwiR hyperparameters for Qwen3-32B are not publicly available, we directly adopt the baseline results reported in~\cite{shi2025swireasoningswitchthinkinglatentexplicit}. All methods use the same decoding settings: temperature 0.6, top-$p$ 0.95, top-$k$ 20, and min-$p$ 0.0. Dataset-specific entropy thresholds for SeLaR are provided in Appendix~\ref{app:Sensitivity Analysis Details}. To further assess generality across model families, we additionally report results on DeepSeek-R1-Distill-Llama-8B~\cite{deepseekai2025deepseekr1incentivizingreasoningcapability} in Appendix~\ref{appdix:Deepseek}. 

\subsection{Results}
Table~\ref{tab:bridge_results} presents the main results across five reasoning benchmarks and three model scales.

\begin{tcolorbox}[colback=Gray, colframe=darkgreen, boxrule=0.5mm]
\textbf{Finding 1:} SeLaR consistently outperforms baselines on average.
\end{tcolorbox}
Across all model scales, SeLaR achieves the highest average accuracy, improving upon CoT (Sampling) by +0.75\%, +3.88\%, and +2.45\% on Qwen3-1.7B, Qwen3-8B, and Qwen3-32B, respectively. Notably, SeLaR is the only method that consistently surpasses CoT across all model sizes. In contrast, Soft Thinking and SwiR exhibit inconsistent behavior: while occasionally matching or exceeding CoT on individual benchmarks, their average performance frequently falls below the CoT baseline. 
\begin{figure*}[t]
\centering
\includegraphics[width=\textwidth]{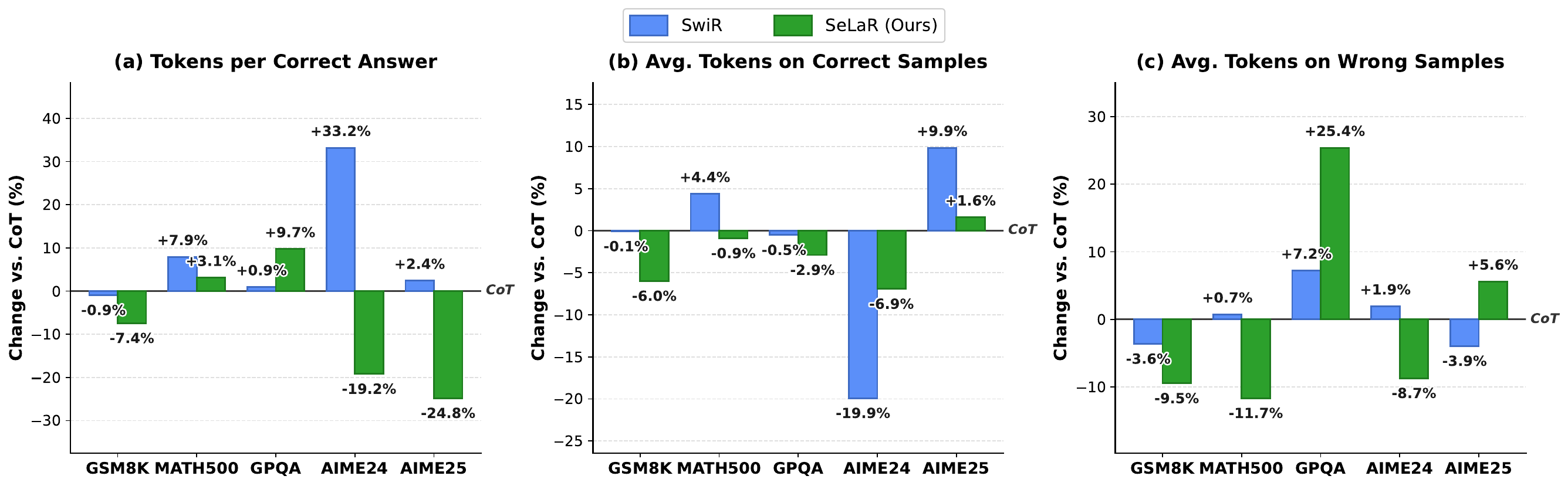}
\caption{%
Computational overhead of SeLaR and SwiR relative to CoT (Sampling) on Qwen3-8B.
\textbf{(a)} Tokens per correct answer: SeLaR beats SwiR on 4 of 5 benchmarks, most notably on AIME~2024/2025.
\textbf{(b)} Average tokens on correctly-answered samples.
\textbf{(c)} Average tokens on wrongly-answered samples.
}
\label{fig:overhead}
\end{figure*}
\begin{tcolorbox}[colback=Gray, colframe=darkgreen, boxrule=0.5mm]
\textbf{Finding 2:} Significant gains on challenging benchmarks.
\end{tcolorbox}
SeLaR's advantage is most pronounced on the hardest benchmarks, AIME 2024 and AIME 2025, which demand deep multi-step reasoning and precise numerical computation. On Qwen3-8B, SeLaR improves AIME 2024 from 76.67\% to 83.33\% (+6.66\%) and AIME 2025 from 66.67\% to 80.00\% (+13.33\%). Similar trends hold on Qwen3-1.7B and Qwen3-32B. We attribute these gains to the complementary effect of the two components: entropy gating concentrates latent reasoning at the most uncertain and consequential steps, while contrastive regularization prevents premature collapse at precisely those steps where a wrong commitment would cascade into irreversible reasoning errors. Together, they provide the greatest benefit on problems where a single misstep is most costly.

\subsection{Computational Overhead}
\label{sec:overhead}

The cost-effectiveness of a reasoning method is jointly determined by the tokens it spends per problem and the accuracy it achieves. Reporting either in isolation is misleading: a method that reduces token usage while sacrificing accuracy is not truly more efficient, and one that improves accuracy at disproportionate token cost is not truly faster. We therefore report three complementary metrics that together characterize cost-effectiveness in Figure~\ref{fig:overhead}, all as percentage changes relative to the CoT (Sampling) baseline on Qwen3-8B. 

We denote the average tokens on correctly- and wrongly-answered samples as $T_c$ and $T_w$, respectively, and write accuracy as $\alpha$. Our headline metric is \textbf{\textit{Tokens per Correct Answer}}:
\begin{equation}
\textit{TPCA} = \frac{\alpha \cdot T_c + (1-\alpha) \cdot T_w}{\alpha},
\label{eq:tpca}
\end{equation}

\begin{tcolorbox}[colback=Gray, colframe=darkgreen, boxrule=0.5mm]
\textbf{Finding 3:} On average, SeLaR is more cost-effective than SwiR, with the advantage widening on the hardest reasoning tasks.
\end{tcolorbox}
On TPCA (Figure~\ref{fig:overhead}a), SeLaR outperforms SwiR by $6.5$, $4.8$, $52.4$, and $27.2$ percentage points on GSM8K, MATH500, AIME~2024, and AIME~2025, respectively. This advantage is most pronounced on AIME~2024, where SeLaR reduces TPCA by $19.2\%$ relative to CoT while SwiR inflates it by $33.2\%$. The main reason is that SwiR's accuracy on AIME~2024 drops to $60\%$, only marginally compensated by its shorter reasoning trajectories. The sole exception is GPQA, where SeLaR's TPCA is $9.7\%$ higher than CoT. We attribute this to GPQA's knowledge-intensive nature, where latent exploration at uncertain decoding steps provides little benefit when correct answers depend on domain recall rather than multi-step reasoning.

\begin{tcolorbox}[colback=Gray, colframe=darkgreen, boxrule=0.5mm]
\textbf{Finding 4:} SwiR's apparent efficiency advantage on correct samples is a survivorship bias.
\end{tcolorbox}
SeLaR's gains arise from both shorter correct trajectories and higher accuracy. On correctly-answered samples (Figure~\ref{fig:overhead}b), SeLaR's $T_c$ falls within $-6.0\%$ to $+1.6\%$ of CoT across all benchmarks, confirming that selective activation and contrastive regularization introduce no runtime overhead. However, SwiR's apparent $-19.9\%$ reduction on AIME~2024 is an artifact of survivorship bias, as SwiR answers only the easier $60\%$ of problems correctly, and those naturally require fewer tokens. TPCA corrects for this survivorship bias, revealing SwiR's true $+33.2\%$ cost inflation on AIME~2024.

\begin{figure*}[t]
\centering
\includegraphics[width=\textwidth]{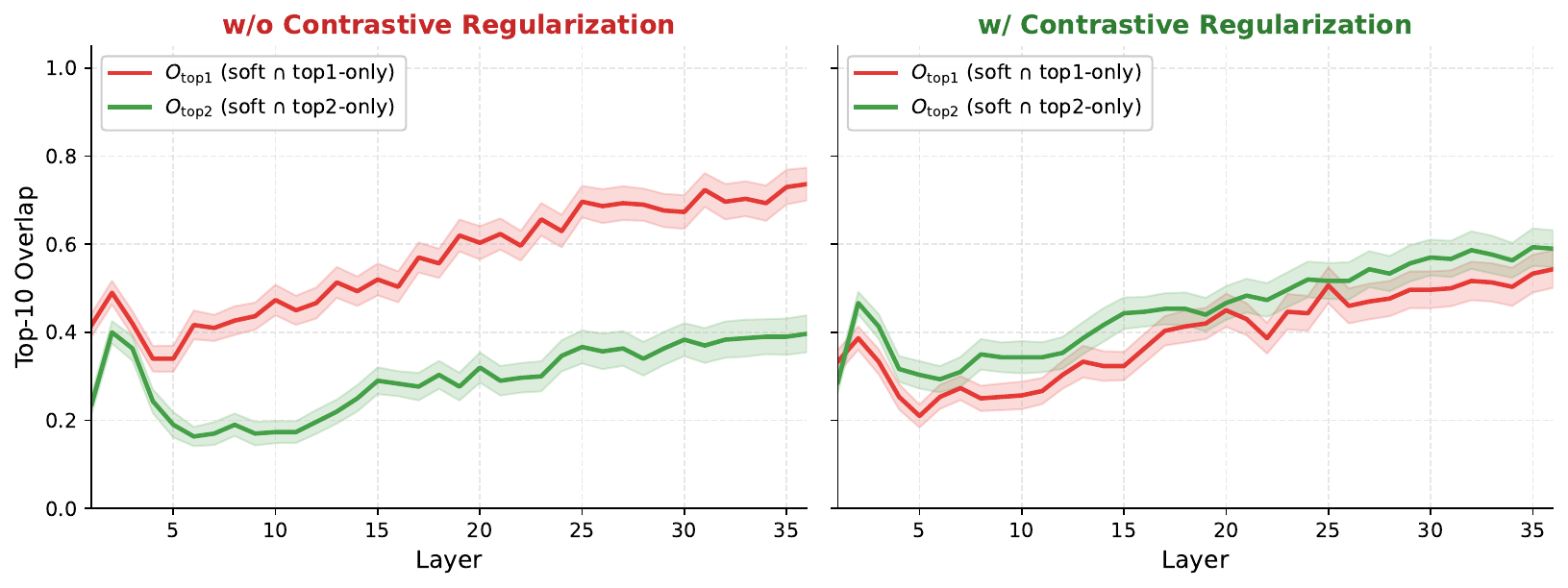}
\caption{%
Layer-wise top-$k$ overlap ($k{=}10$) between the soft-embedding forward pass and single-token reference passes, aggregated over $N{=}200$ branching steps from AIME~2025 on Qwen3-8B. Shaded bands denote standard error.
\textbf{Left:} without contrastive regularization, $O_{\text{top1}}$ dominates $O_{\text{top2}}$ in deep layers, reproducing the collapse behavior of \cite{wu2025llmssinglethreadedreasonersdemystifying}.
\textbf{Right:} with contrastive regularization, \emph{both} overlaps remain non-zero and comparable, indicating multiple reasoning trajectories coexist rather than a single one being swapped.%
}
\label{fig:overlap_aggregate}
\end{figure*}

\subsection{Ablation Studies}

We ablate each component of SeLaR on Qwen3-8B to quantify its individual contribution. Results are presented in Table~\ref{tab:ablation}.
\paragraph{Effect of Selective Activation}
Removing selective activation (i.e., applying soft embeddings globally at every step) leads to an average performance drop of 5.19\% (from 83.56\% to 78.37\%), even falling below the CoT baseline. This confirms that indiscriminate activation perturbs high-confidence steps where the model is already committed, destabilizing reasoning chains that would otherwise succeed.

\paragraph{Effect of Contrastive Regularization}
Removing contrastive regularization results in an average performance drop of $7.82\%$ (from $83.56\%$ to $75.74\%$). The degradation is particularly severe on challenging benchmarks: AIME~2024 drops from $83.33\%$ to $70.00\%$ and AIME~2025 drops from $80.00\%$ to $60.00\%$. While this behavioral evidence confirms that contrastive regularization is indispensable, it does not reveal \emph{how} the component intervenes inside the forward pass. We investigate this mechanism in the following subsection.

\subsection{How Contrastive Regularization Works}

To understand \emph{how} contrastive regularization works, we must answer two questions: (i) does it produce consistent effects across exploratory steps rather than isolated cases, and (ii) does it truly preserve multiple reasoning trajectories, or does it merely swap the dominant top-1 token for a different top-$k$ candidate while remaining single-threaded? We address both via a logit lens analysis inspired by the approach of \cite{nostalgebraist2020logitlens}.

\paragraph{Logit Lens Setting.}
We collect $N=200$ branching steps from 10 random AIME~2025 problems under SeLaR on Qwen3-8B, where a branching step is an exploratory step (entropy above $\tau$) with top-1/top-2 probability ratio below $2.0$. For each step $t$, we cache the KV state and run four independent forward passes at step $t{+}1$, with input embedding set to $e_{v^*_t}$, $e_{v^{**}_t}$, $e_t$, and $\tilde e_t$ respectively, where $v^*_t, v^{**}_t$ are the top-1/top-2 tokens and $e_t, \tilde e_t$ are the soft embeddings without and with contrastive regularization. For each pass, we apply the logit lens at every layer $\ell$ and take the top-$k$ projected tokens ($k{=}10$), denoted $\mathcal{T}^{\,\cdot}_{\ell}$. We then measure how much each soft-embedding pass shares with the two references:
\begin{equation}
O_{\text{top1}}(\ell) = \frac{|\mathcal{T}^{\text{soft}}_{\ell} \cap \mathcal{T}^{\text{top1}}_{\ell}|}{k},
\end{equation}
\begin{equation}
O_{\text{top2}}(\ell) = \frac{|\mathcal{T}^{\text{soft}}_{\ell} \cap \mathcal{T}^{\text{top2}}_{\ell}|}{k},
\label{eq:overlap}
\end{equation}
quantifying how much of each candidate's reasoning content the soft-embedding forward pass still carries at layer $\ell$. We average both across all $N$ steps.

\begin{table*}[ht]
\centering
\caption{Ablation study on Qwen3-8B. We evaluate the contribution of each component in SeLaR.}
\label{tab:ablation}
\begin{tabular}{l|ccccc|c}
\toprule
Method & GSM8K & MATH500 & GPQA & AIME24 & AIME25 & Avg \\
\midrule
CoT (Sampling) & 95.45 & \textbf{98.00} & 61.62 & 76.67 & 66.67 & 79.68 \\
\midrule
\rowcolor{lightgreen} \textbf{SeLaR (Full)} & \textbf{95.83} & 97.00 & \textbf{61.62} & \textbf{83.33} & \textbf{80.00} & \textbf{83.56} \\
\midrule
\multicolumn{7}{l}{\textit{Component Ablation}} \\
\quad w/o Selective Activation & 95.14 & 95.80 & 57.58 & 76.67 & 66.67 & 78.37 \\
\quad w/o Contrastive Reg. & 94.92 & 96.20 & 57.58 & 70.00 & 60.00 & 75.74 \\
\bottomrule
\end{tabular}
\end{table*}

\paragraph{Logit Lens Results.}
Figure~\ref{fig:overlap_aggregate} shows the aggregated curves. Without contrastive regularization (left), $O_{\text{top1}}$ rises from ${\sim}0.45$ to ${\sim}0.73$ while $O_{\text{top2}}$ stagnates around ${\sim}0.40$, reproducing the collapse behavior of \cite{wu2025llmssinglethreadedreasonersdemystifying}: the forward pass progressively collapses onto the top-1 trajectory and suppresses the top-2 alternative. Unlike \cite{wu2025llmssinglethreadedreasonersdemystifying}, $O_{\text{top1}}$ does not saturate to $1$ because our soft embedding mixes $k$ top candidates computed from the model's actual output distribution, rather than a manually balanced two-token mixture. The key observation is thus the relative gap between the two curves, not their absolute values.

\begin{tcolorbox}[colback=Gray, colframe=darkgreen, boxrule=0.5mm]
\textbf{Finding 5:} Contrastive regularization prevents the collapse into single-threaded behavior, keeping multiple candidate trajectories alive in deep layers.
\end{tcolorbox}

A natural concern is whether contrastive regularization merely shifts the collapse from the top-1 to another top-$k$ candidate, leaving the forward pass still single-threaded. Note that if this were the case, we would expect $O_{\text{top1}}$ to decrease and $O_{\text{top2}}$ to increase, mirroring the left panel with the two curves swapped. Instead, both overlaps remain substantially non-zero and comparable in deep layers: $O_{\text{top2}}$ climbs to ${\sim}0.60$ while $O_{\text{top1}}$ settles at ${\sim}0.55$. This confirms that contrastive regularization implements genuine \textbf{\textit{sustained exploration of latent reasoning paths}}.

\begin{figure}[ht]
\centering
\includegraphics[width=\columnwidth]{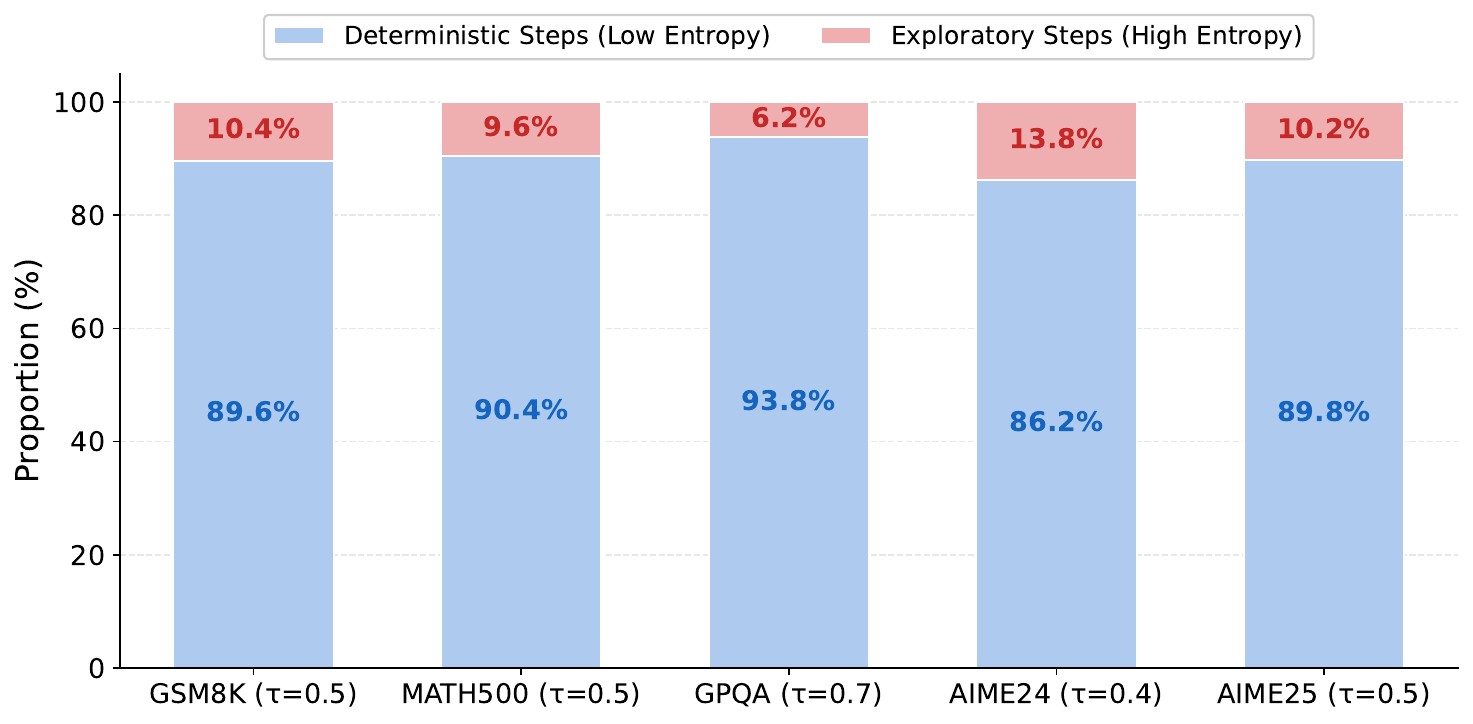}
\caption{Activation frequency analysis on Qwen3-8B. Exploratory steps (high entropy) account for 6.2\%--13.8\% of total reasoning tokens.}
\label{fig:activation_frequency}
\end{figure}

\subsection{Detailed Analysis}
\paragraph{Sensitivity Analysis}
\label{par:Sensitivity Analysis}
Appendix~\ref{app:Sensitivity Analysis Details} examines the sensitivity of SeLaR to the entropy threshold $\tau$ and top-$k$ value. We find that performance remains stable across $\tau \in [0.3, 0.7]$ with $k=3$, achieving the best average accuracy (80.86\%) at $\tau=0.5$. For the top-$k$ value, $k=3$ consistently outperforms larger values, as excessive candidates dilute the soft embedding with low-probability tokens.

\paragraph{Activation Frequency Analysis}
Figure~\ref{fig:activation_frequency} shows the proportion of exploratory steps (high entropy) versus deterministic steps (low entropy) across benchmarks. Exploratory steps account for 6.2\%--13.8\% of total tokens, averaging 10.0\%. This variation reflects dataset-specific thresholds: AIME 2024 ($\tau=0.4$) exhibits the highest activation frequency, while GPQA-Diamond ($\tau=0.7$) shows the lowest. This confirms that selective activation targets approximately one in ten tokens where exploration is most beneficial. Activation frequency analysis on DeepSeek-R1-Distill-Llama-8B is provided in Appendix~\ref{appdix:Deepseek}.

\section{Conclusion}
We present \textbf{SeLaR}, a training-free latent reasoning framework that selectively activates soft embeddings based on entropy. SeLaR preserves discrete token commitments at deterministic steps for stability, and activates soft embeddings at exploratory steps to enable alternative reasoning trajectories. An entropy-aware contrastive regularization further mitigates premature collapse toward the dominant token. More broadly, SeLaR addresses both \emph{when} and \emph{how} latent reasoning should be applied: token-level entropy signals when to activate, while contrastive regularization governs how exploration is sustained. This perspective offers new insights into designing adaptive reasoning mechanisms in large language models.

\section*{Acknowledgments}
This work is supported by National Key Research and Development Program of China (2024YFE0203100),
Guangdong Provincial Key Laboratory of Ultra High Definition Immersive Media Technology (Grant No.
2024B1212010006), and Shenzhen Science and Technology Program (JCYJ20230807120800001).

\clearpage
\section*{Limitations}
While SeLaR demonstrates consistent improvements across multiple benchmarks, some limitations warrant discussion.

\paragraph{Constraints of Token Embedding Space}
Like other training-free latent reasoning methods, SeLaR operates in the token embedding space at the input level. Although contrastive regularization effectively mitigates premature collapse toward the dominant token, this approach is inherently limited in expressiveness compared to manipulating hidden states directly. Future work on latent reasoning should explore the hidden state space, which serves as the primary information carrier for reasoning in LLMs.

\paragraph{Sensitivity to Base Model Confidence}
SeLaR yields larger improvements on base models with higher confidence (e.g., Qwen3-8B) than on those with lower confidence (e.g., DeepSeek-R1-Distill-Llama-8B). Our analysis indicates that less confident models exhibit higher entropy more frequently, triggering excessive exploratory steps. Future work should investigate confidence-aware activation mechanisms or explore signals beyond entropy to better adapt to varying base model characteristics.

\bibliography{main}

\begin{thebibliography}{57}
\providecommand{\natexlab}[1]{#1}

\bibitem[{Abdin et~al.(2025)Abdin, Agarwal, Awadallah, Balachandran, Behl, Chen, de~Rosa, Gunasekar, Javaheripi, Joshi, Kauffmann, Lara, Mendes, Mitra, Nushi, Papailiopoulos, Saarikivi, Shah, Shrivastava, Vineet, Wu, Yousefi, and Zheng}]{abdin2025phi4reasoningtechnicalreport}
Marah Abdin, Sahaj Agarwal, Ahmed Awadallah, Vidhisha Balachandran, Harkirat Behl, Lingjiao Chen, Gustavo de~Rosa, Suriya Gunasekar, Mojan Javaheripi, Neel Joshi, Piero Kauffmann, Yash Lara, Caio César~Teodoro Mendes, Arindam Mitra, Besmira Nushi, Dimitris Papailiopoulos, Olli Saarikivi, Shital Shah, Vaishnavi Shrivastava, and 4 others. 2025.
\newblock \href {https://arxiv.org/abs/2504.21318} {Phi-4-reasoning technical report}.
\newblock \emph{Preprint}, arXiv:2504.21318.

\bibitem[{Besta et~al.(2024)Besta, Blach, Kubicek, Gerstenberger, Podstawski, Gianinazzi, Gajda, Lehmann, Niewiadomski, Nyczyk, and Hoefler}]{Besta_2024}
Maciej Besta, Nils Blach, Ales Kubicek, Robert Gerstenberger, Michal Podstawski, Lukas Gianinazzi, Joanna Gajda, Tomasz Lehmann, Hubert Niewiadomski, Piotr Nyczyk, and Torsten Hoefler. 2024.
\newblock \href {https://doi.org/10.1609/aaai.v38i16.29720} {Graph of thoughts: Solving elaborate problems with large language models}.
\newblock \emph{Proceedings of the AAAI Conference on Artificial Intelligence}, 38(16):17682–17690.

\bibitem[{Brown et~al.(2020)Brown, Mann, Ryder, Subbiah, Kaplan, Dhariwal, Neelakantan, Shyam, Sastry, Askell, Agarwal, Herbert-Voss, Krueger, Henighan, Child, Ramesh, Ziegler, Wu, Winter, Hesse, Chen, Sigler, Litwin, Gray, Chess, Clark, Berner, McCandlish, Radford, Sutskever, and Amodei}]{brown2020languagemodelsfewshotlearners}
Tom~B. Brown, Benjamin Mann, Nick Ryder, Melanie Subbiah, Jared Kaplan, Prafulla Dhariwal, Arvind Neelakantan, Pranav Shyam, Girish Sastry, Amanda Askell, Sandhini Agarwal, Ariel Herbert-Voss, Gretchen Krueger, Tom Henighan, Rewon Child, Aditya Ramesh, Daniel~M. Ziegler, Jeffrey Wu, Clemens Winter, and 12 others. 2020.
\newblock \href {https://arxiv.org/abs/2005.14165} {Language models are few-shot learners}.
\newblock \emph{Preprint}, arXiv:2005.14165.

\bibitem[{Cheng and Durme(2024)}]{cheng2024compressedchainthoughtefficient}
Jeffrey Cheng and Benjamin~Van Durme. 2024.
\newblock \href {https://arxiv.org/abs/2412.13171} {Compressed chain of thought: Efficient reasoning through dense representations}.
\newblock \emph{Preprint}, arXiv:2412.13171.

\bibitem[{Chowdhery et~al.(2022)Chowdhery, Narang, Devlin, Bosma, Mishra, Roberts, Barham, Chung, Sutton, Gehrmann, Schuh, Shi, Tsvyashchenko, Maynez, Rao, Barnes, Tay, Shazeer, Prabhakaran, Reif, Du, Hutchinson, Pope, Bradbury, Austin, Isard, Gur-Ari, Yin, Duke, Levskaya, Ghemawat, Dev, Michalewski, Garcia, Misra, Robinson, Fedus, Zhou, Ippolito, Luan, Lim, Zoph, Spiridonov, Sepassi, Dohan, Agrawal, Omernick, Dai, Pillai, Pellat, Lewkowycz, Moreira, Child, Polozov, Lee, Zhou, Wang, Saeta, Diaz, Firat, Catasta, Wei, Meier-Hellstern, Eck, Dean, Petrov, and Fiedel}]{chowdhery2022palmscalinglanguagemodeling}
Aakanksha Chowdhery, Sharan Narang, Jacob Devlin, Maarten Bosma, Gaurav Mishra, Adam Roberts, Paul Barham, Hyung~Won Chung, Charles Sutton, Sebastian Gehrmann, Parker Schuh, Kensen Shi, Sasha Tsvyashchenko, Joshua Maynez, Abhishek Rao, Parker Barnes, Yi~Tay, Noam Shazeer, Vinodkumar Prabhakaran, and 48 others. 2022.
\newblock \href {https://arxiv.org/abs/2204.02311} {Palm: Scaling language modeling with pathways}.
\newblock \emph{Preprint}, arXiv:2204.02311.

\bibitem[{Chu et~al.(2024)Chu, Chen, Chen, Yu, He, Wang, Peng, Liu, Qin, and Liu}]{chu2024navigateenigmaticlabyrinthsurvey}
Zheng Chu, Jingchang Chen, Qianglong Chen, Weijiang Yu, Tao He, Haotian Wang, Weihua Peng, Ming Liu, Bing Qin, and Ting Liu. 2024.
\newblock \href {https://arxiv.org/abs/2309.15402} {Navigate through enigmatic labyrinth a survey of chain of thought reasoning: Advances, frontiers and future}.
\newblock \emph{Preprint}, arXiv:2309.15402.

\bibitem[{DeepSeek-AI et~al.(2025)DeepSeek-AI, Guo, Yang, Zhang, Song, Zhang, Xu, Zhu, Ma, Wang, Bi, Zhang, Yu, Wu, Wu, Gou, Shao, Li, Gao, Liu, Xue, Wang, Wu, Feng, Lu, Zhao, Deng, Zhang, Ruan, Dai, Chen, Ji, Li, Lin, Dai, Luo, Hao, Chen, Li, Zhang, Bao, Xu, Wang, Ding, Xin, Gao, Qu, Li, Guo, Li, Wang, Chen, Yuan, Qiu, Li, Cai, Ni, Liang, Chen, Dong, Hu, Gao, Guan, Huang, Yu, Wang, Zhang, Zhao, Wang, Zhang, Xu, Xia, Zhang, Zhang, Tang, Li, Wang, Li, Tian, Huang, Zhang, Wang, Chen, Du, Ge, Zhang, Pan, Wang, Chen, Jin, Chen, Lu, Zhou, Chen, Ye, Wang, Yu, Zhou, Pan, Li, Zhou, Wu, Ye, Yun, Pei, Sun, Wang, Zeng, Zhao, Liu, Liang, Gao, Yu, Zhang, Xiao, An, Liu, Wang, Chen, Nie, Cheng, Liu, Xie, Liu, Yang, Li, Su, Lin, Li, Jin, Shen, Chen, Sun, Wang, Song, Zhou, Wang, Shan, Li, Wang, Wei, Zhang, Xu, Li, Zhao, Sun, Wang, Yu, Zhang, Shi, Xiong, He, Piao, Wang, Tan, Ma, Liu, Guo, Ou, Wang, Gong, Zou, He, Xiong, Luo, You, Liu, Zhou, Zhu, Xu, Huang, Li, Zheng, Zhu, Ma, Tang, Zha, Yan, Ren, Ren, Sha, Fu, Xu, Xie, Zhang,
  Hao, Ma, Yan, Wu, Gu, Zhu, Liu, Li, Xie, Song, Pan, Huang, Xu, Zhang, and Zhang}]{deepseekai2025deepseekr1incentivizingreasoningcapability}
DeepSeek-AI, Daya Guo, Dejian Yang, Haowei Zhang, Junxiao Song, Ruoyu Zhang, Runxin Xu, Qihao Zhu, Shirong Ma, Peiyi Wang, Xiao Bi, Xiaokang Zhang, Xingkai Yu, Yu~Wu, Z.~F. Wu, Zhibin Gou, Zhihong Shao, Zhuoshu Li, Ziyi Gao, and 181 others. 2025.
\newblock \href {https://arxiv.org/abs/2501.12948} {Deepseek-r1: Incentivizing reasoning capability in llms via reinforcement learning}.
\newblock \emph{Preprint}, arXiv:2501.12948.

\bibitem[{Deng et~al.(2023)Deng, Prasad, Fernandez, Smolensky, Chaudhary, and Shieber}]{deng2023implicitchainthoughtreasoning}
Yuntian Deng, Kiran Prasad, Roland Fernandez, Paul Smolensky, Vishrav Chaudhary, and Stuart Shieber. 2023.
\newblock \href {https://arxiv.org/abs/2311.01460} {Implicit chain of thought reasoning via knowledge distillation}.
\newblock \emph{Preprint}, arXiv:2311.01460.

\bibitem[{Du et~al.(2022)Du, Qian, Liu, Ding, Qiu, Yang, and Tang}]{du2022glmgenerallanguagemodel}
Zhengxiao Du, Yujie Qian, Xiao Liu, Ming Ding, Jiezhong Qiu, Zhilin Yang, and Jie Tang. 2022.
\newblock \href {https://arxiv.org/abs/2103.10360} {Glm: General language model pretraining with autoregressive blank infilling}.
\newblock \emph{Preprint}, arXiv:2103.10360.

\bibitem[{Geiping et~al.(2025)Geiping, McLeish, Jain, Kirchenbauer, Singh, Bartoldson, Kailkhura, Bhatele, and Goldstein}]{geiping2025scalingtesttimecomputelatent}
Jonas Geiping, Sean McLeish, Neel Jain, John Kirchenbauer, Siddharth Singh, Brian~R. Bartoldson, Bhavya Kailkhura, Abhinav Bhatele, and Tom Goldstein. 2025.
\newblock \href {https://arxiv.org/abs/2502.05171} {Scaling up test-time compute with latent reasoning: A recurrent depth approach}.
\newblock \emph{Preprint}, arXiv:2502.05171.

\bibitem[{Goyal et~al.(2024)Goyal, Ji, Rawat, Menon, Kumar, and Nagarajan}]{goyal2024thinkspeaktraininglanguage}
Sachin Goyal, Ziwei Ji, Ankit~Singh Rawat, Aditya~Krishna Menon, Sanjiv Kumar, and Vaishnavh Nagarajan. 2024.
\newblock \href {https://arxiv.org/abs/2310.02226} {Think before you speak: Training language models with pause tokens}.
\newblock \emph{Preprint}, arXiv:2310.02226.

\bibitem[{Hao et~al.(2025)Hao, Sukhbaatar, Su, Li, Hu, Weston, and Tian}]{hao2025traininglargelanguagemodels}
Shibo Hao, Sainbayar Sukhbaatar, DiJia Su, Xian Li, Zhiting Hu, Jason Weston, and Yuandong Tian. 2025.
\newblock \href {https://arxiv.org/abs/2412.06769} {Training large language models to reason in a continuous latent space}.
\newblock \emph{Preprint}, arXiv:2412.06769.

\bibitem[{Havrilla et~al.(2024)Havrilla, Du, Raparthy, Nalmpantis, Dwivedi-Yu, Zhuravinskyi, Hambro, Sukhbaatar, and Raileanu}]{havrilla2024teachinglargelanguagemodels}
Alex Havrilla, Yuqing Du, Sharath~Chandra Raparthy, Christoforos Nalmpantis, Jane Dwivedi-Yu, Maksym Zhuravinskyi, Eric Hambro, Sainbayar Sukhbaatar, and Roberta Raileanu. 2024.
\newblock \href {https://arxiv.org/abs/2403.04642} {Teaching large language models to reason with reinforcement learning}.
\newblock \emph{Preprint}, arXiv:2403.04642.

\bibitem[{Hendrycks et~al.(2021)Hendrycks, Burns, Kadavath, Arora, Basart, Tang, Song, and Steinhardt}]{hendrycks2021measuringmathematicalproblemsolving}
Dan Hendrycks, Collin Burns, Saurav Kadavath, Akul Arora, Steven Basart, Eric Tang, Dawn Song, and Jacob Steinhardt. 2021.
\newblock \href {https://arxiv.org/abs/2103.03874} {Measuring mathematical problem solving with the math dataset}.
\newblock \emph{Preprint}, arXiv:2103.03874.

\bibitem[{{HuggingFaceH4}(2024)}]{huggingfaceh42024aime}
{HuggingFaceH4}. 2024.
\newblock \href {https://huggingface.co/datasets/HuggingFaceH4/aime_2024} {{AIME 2024: American Invitational Mathematics Examination}}.
\newblock Hugging Face Dataset.

\bibitem[{Jin et~al.(2025)Jin, Luo, Cheng, Wang, Hua, Tang, Wang, and Zhang}]{jin2025disentanglingmemoryreasoningability}
Mingyu Jin, Weidi Luo, Sitao Cheng, Xinyi Wang, Wenyue Hua, Ruixiang Tang, William~Yang Wang, and Yongfeng Zhang. 2025.
\newblock \href {https://arxiv.org/abs/2411.13504} {Disentangling memory and reasoning ability in large language models}.
\newblock \emph{Preprint}, arXiv:2411.13504.

\bibitem[{Lee et~al.(2025)Lee, Fischer, Wu, Marwood, Baluja, Schuurmans, and Chen}]{lee2025evolvingdeeperllmthinking}
Kuang-Huei Lee, Ian Fischer, Yueh-Hua Wu, Dave Marwood, Shumeet Baluja, Dale Schuurmans, and Xinyun Chen. 2025.
\newblock \href {https://arxiv.org/abs/2501.09891} {Evolving deeper llm thinking}.
\newblock \emph{Preprint}, arXiv:2501.09891.

\bibitem[{Li et~al.(2025)Li, Fu, Fan, Liu, Shu, Qin, Yang, King, and Ying}]{li2025implicitreasoninglargelanguage}
Jindong Li, Yali Fu, Li~Fan, Jiahong Liu, Yao Shu, Chengwei Qin, Menglin Yang, Irwin King, and Rex Ying. 2025.
\newblock \href {https://arxiv.org/abs/2509.02350} {Implicit reasoning in large language models: A comprehensive survey}.
\newblock \emph{Preprint}, arXiv:2509.02350.

\bibitem[{Lobo et~al.(2025)Lobo, Agarwal, and Lakkaraju}]{lobo2025impactfinetuningchainofthoughtreasoning}
Elita Lobo, Chirag Agarwal, and Himabindu Lakkaraju. 2025.
\newblock \href {https://arxiv.org/abs/2411.15382} {On the impact of fine-tuning on chain-of-thought reasoning}.
\newblock \emph{Preprint}, arXiv:2411.15382.

\bibitem[{Madaan et~al.(2023)Madaan, Tandon, Gupta, Hallinan, Gao, Wiegreffe, Alon, Dziri, Prabhumoye, Yang, Gupta, Majumder, Hermann, Welleck, Yazdanbakhsh, and Clark}]{madaan2023selfrefineiterativerefinementselffeedback}
Aman Madaan, Niket Tandon, Prakhar Gupta, Skyler Hallinan, Luyu Gao, Sarah Wiegreffe, Uri Alon, Nouha Dziri, Shrimai Prabhumoye, Yiming Yang, Shashank Gupta, Bodhisattwa~Prasad Majumder, Katherine Hermann, Sean Welleck, Amir Yazdanbakhsh, and Peter Clark. 2023.
\newblock \href {https://arxiv.org/abs/2303.17651} {Self-refine: Iterative refinement with self-feedback}.
\newblock \emph{Preprint}, arXiv:2303.17651.

\bibitem[{Mohtashami et~al.(2024)Mohtashami, Pagliardini, and Jaggi}]{mohtashami2024cotformerchainofthoughtdrivenarchitecture}
Amirkeivan Mohtashami, Matteo Pagliardini, and Martin Jaggi. 2024.
\newblock \href {https://arxiv.org/abs/2310.10845} {Cotformer: A chain-of-thought driven architecture with budget-adaptive computation cost at inference}.
\newblock \emph{Preprint}, arXiv:2310.10845.

\bibitem[{Nostalgebraist(2020)}]{nostalgebraist2020logitlens}
Nostalgebraist. 2020.
\newblock \href {https://www.lesswrong.com/posts/AcKRB8wDpdaN6v6ru/interpreting-gpt-the-logit-lens} {Interpreting gpt: The logit lens}.
\newblock Blog post on LessWrong.

\bibitem[{OpenAI et~al.(2025)OpenAI, :, Agarwal, Ahmad, Ai, Altman, Applebaum, Arbus, Arora, Bai, Baker, Bao, Barak, Bennett, Bertao, Brett, Brevdo, Brockman, Bubeck, Chang, Chen, Chen, Cheung, Clark, Cook, Dukhan, Dvorak, Fives, Fomenko, Garipov, Georgiev, Glaese, Gogineni, Goucher, Gross, Guzman, Hallman, Hehir, Heidecke, Helyar, Hu, Huet, Huh, Jain, Johnson, Koch, Kofman, Kundel, Kwon, Kyrylov, Le, Leclerc, Lennon, Lessans, Lezcano-Casado, Li, Li, Lin, Liss, Lily, Liu, Liu, Lu, Lu, Martinovic, McCallum, McGrath, McKinney, McLaughlin, Mei, Mostovoy, Mu, Myles, Neitz, Nichol, Pachocki, Paino, Palmie, Pantuliano, Parascandolo, Park, Pathak, Paz, Peran, Pimenov, Pokrass, Proehl, Qiu, Raila, Raso, Ren, Richardson, Robinson, Rotsted, Salman, Sanjeev, Schwarzer, Sculley, Sikchi, Simon, Singhal, Song, Stuckey, Sun, Tillet, Toizer, Tsimpourlas, Vyas, Wallace, Wang, Wang, Watkins, Weil, Wendling, Whinnery, Whitney, Wong, Yang, Yang, Yasunaga, Ying, Zaremba, Zhan, Zhang, Zhang, Zhang, and
  Zhao}]{openai2025gptoss120bgptoss20bmodel}
OpenAI, :, Sandhini Agarwal, Lama Ahmad, Jason Ai, Sam Altman, Andy Applebaum, Edwin Arbus, Rahul~K. Arora, Yu~Bai, Bowen Baker, Haiming Bao, Boaz Barak, Ally Bennett, Tyler Bertao, Nivedita Brett, Eugene Brevdo, Greg Brockman, Sebastien Bubeck, and 108 others. 2025.
\newblock \href {https://arxiv.org/abs/2508.10925} {gpt-oss-120b \& gpt-oss-20b model card}.
\newblock \emph{Preprint}, arXiv:2508.10925.

\bibitem[{OpenAI et~al.(2024{\natexlab{a}})OpenAI, :, Jaech, Kalai, Lerer, Richardson, El-Kishky, Low, Helyar, Madry, Beutel, Carney, Iftimie, Karpenko, Passos, Neitz, Prokofiev, Wei, Tam, Bennett, Kumar, Saraiva, Vallone, Duberstein, Kondrich, Mishchenko, Applebaum, Jiang, Nair, Zoph, Ghorbani, Rossen, Sokolowsky, Barak, McGrew, Minaiev, Hao, Baker, Houghton, McKinzie, Eastman, Lugaresi, Bassin, Hudson, Li, de~Bourcy, Voss, Shen, Zhang, Koch, Orsinger, Hesse, Fischer, Chan, Roberts, Kappler, Levy, Selsam, Dohan, Farhi, Mely, Robinson, Tsipras, Li, Oprica, Freeman, Zhang, Wong, Proehl, Cheung, Mitchell, Wallace, Ritter, Mays, Wang, Such, Raso, Leoni, Tsimpourlas, Song, von Lohmann, Sulit, Salmon, Parascandolo, Chabot, Zhao, Brockman, Leclerc, Salman, Bao, Sheng, Andrin, Bagherinezhad, Ren, Lightman, Chung, Kivlichan, O'Connell, Osband, Gilaberte, Akkaya, Kostrikov, Sutskever, Kofman, Pachocki, Lennon, Wei, Harb, Twore, Feng, Yu, Weng, Tang, Yu, Candela, Palermo, Parish, Heidecke, Hallman, Rizzo, Gordon,
  Uesato, Ward, Huizinga, Wang, Chen, Xiao, Singhal, Nguyen, Cobbe, Shi, Wood, Rimbach, Gu-Lemberg, Liu, Lu, Stone, Yu, Ahmad, Yang, Liu, Maksin, Ho, Fedus, Weng, Li, McCallum, Held, Kuhn, Kondraciuk, Kaiser, Metz, Boyd, Trebacz, Joglekar, Chen, Tintor, Meyer, Jones, Kaufer, Schwarzer, Shah, Yatbaz, Guan, Xu, Yan, Glaese, Chen, Lampe, Malek, Wang, Fradin, McClay, Pavlov, Wang, Wang, Murati, Bavarian, Rohaninejad, McAleese, Chowdhury, Chowdhury, Ryder, Tezak, Brown, Nachum, Boiko, Murk, Watkins, Chao, Ashbourne, Izmailov, Zhokhov, Dias, Arora, Lin, Lopes, Gaon, Miyara, Leike, Hwang, Garg, Brown, James, Shu, Cheu, Greene, Jain, Altman, Toizer, Toyer, Miserendino, Agarwal, Hernandez, Baker, McKinney, Yan, Zhao, Hu, Santurkar, Chaudhuri, Zhang, Fu, Papay, Lin, Balaji, Sanjeev, Sidor, Broda, Clark, Wang, Gordon, Sanders, Patwardhan, Sottiaux, Degry, Dimson, Zheng, Garipov, Stasi, Bansal, Creech, Peterson, Eloundou, Qi, Kosaraju, Monaco, Pong, Fomenko, Zheng, Zhou, McCabe, Zaremba, Dubois, Lu, Chen, Cha, Bai, He,
  Zhang, Wang, Shao, and Li}]{openai2024openaio1card}
OpenAI, :, Aaron Jaech, Adam Kalai, Adam Lerer, Adam Richardson, Ahmed El-Kishky, Aiden Low, Alec Helyar, Aleksander Madry, Alex Beutel, Alex Carney, Alex Iftimie, Alex Karpenko, Alex~Tachard Passos, Alexander Neitz, Alexander Prokofiev, Alexander Wei, Allison Tam, and 244 others. 2024{\natexlab{a}}.
\newblock \href {https://arxiv.org/abs/2412.16720} {Openai o1 system card}.
\newblock \emph{Preprint}, arXiv:2412.16720.

\bibitem[{OpenAI et~al.(2024{\natexlab{b}})OpenAI, Achiam, Adler, Agarwal, Ahmad, Akkaya, Aleman, Almeida, Altenschmidt, Altman, Anadkat, Avila, Babuschkin, Balaji, Balcom, Baltescu, Bao, Bavarian, Belgum, Bello, Berdine, Bernadett-Shapiro, Berner, Bogdonoff, Boiko, Boyd, Brakman, Brockman, Brooks, Brundage, Button, Cai, Campbell, Cann, Carey, Carlson, Carmichael, Chan, Chang, Chantzis, Chen, Chen, Chen, Chen, Chen, Chess, Cho, Chu, Chung, Cummings, Currier, Dai, Decareaux, Degry, Deutsch, Deville, Dhar, Dohan, Dowling, Dunning, Ecoffet, Eleti, Eloundou, Farhi, Fedus, Felix, Fishman, Forte, Fulford, Gao, Georges, Gibson, Goel, Gogineni, Goh, Gontijo-Lopes, Gordon, Grafstein, Gray, Greene, Gross, Gu, Guo, Hallacy, Han, Harris, He, Heaton, Heidecke, Hesse, Hickey, Hickey, Hoeschele, Houghton, Hsu, Hu, Hu, Huizinga, Jain, Jain, Jang, Jiang, Jiang, Jin, Jin, Jomoto, Jonn, Jun, Kaftan, Łukasz Kaiser, Kamali, Kanitscheider, Keskar, Khan, Kilpatrick, Kim, Kim, Kim, Kirchner, Kiros, Knight, Kokotajlo, Łukasz
  Kondraciuk, Kondrich, Konstantinidis, Kosic, Krueger, Kuo, Lampe, Lan, Lee, Leike, Leung, Levy, Li, Lim, Lin, Lin, Litwin, Lopez, Lowe, Lue, Makanju, Malfacini, Manning, Markov, Markovski, Martin, Mayer, Mayne, McGrew, McKinney, McLeavey, McMillan, McNeil, Medina, Mehta, Menick, Metz, Mishchenko, Mishkin, Monaco, Morikawa, Mossing, Mu, Murati, Murk, Mély, Nair, Nakano, Nayak, Neelakantan, Ngo, Noh, Ouyang, O'Keefe, Pachocki, Paino, Palermo, Pantuliano, Parascandolo, Parish, Parparita, Passos, Pavlov, Peng, Perelman, de~Avila Belbute~Peres, Petrov, de~Oliveira~Pinto, Michael, Pokorny, Pokrass, Pong, Powell, Power, Power, Proehl, Puri, Radford, Rae, Ramesh, Raymond, Real, Rimbach, Ross, Rotsted, Roussez, Ryder, Saltarelli, Sanders, Santurkar, Sastry, Schmidt, Schnurr, Schulman, Selsam, Sheppard, Sherbakov, Shieh, Shoker, Shyam, Sidor, Sigler, Simens, Sitkin, Slama, Sohl, Sokolowsky, Song, Staudacher, Such, Summers, Sutskever, Tang, Tezak, Thompson, Tillet, Tootoonchian, Tseng, Tuggle, Turley, Tworek, Uribe,
  Vallone, Vijayvergiya, Voss, Wainwright, Wang, Wang, Wang, Ward, Wei, Weinmann, Welihinda, Welinder, Weng, Weng, Wiethoff, Willner, Winter, Wolrich, Wong, Workman, Wu, Wu, Wu, Xiao, Xu, Yoo, Yu, Yuan, Zaremba, Zellers, Zhang, Zhang, Zhao, Zheng, Zhuang, Zhuk, and Zoph}]{openai2024gpt4technicalreport}
OpenAI, Josh Achiam, Steven Adler, Sandhini Agarwal, Lama Ahmad, Ilge Akkaya, Florencia~Leoni Aleman, Diogo Almeida, Janko Altenschmidt, Sam Altman, Shyamal Anadkat, Red Avila, Igor Babuschkin, Suchir Balaji, Valerie Balcom, Paul Baltescu, Haiming Bao, Mohammad Bavarian, Jeff Belgum, and 262 others. 2024{\natexlab{b}}.
\newblock \href {https://arxiv.org/abs/2303.08774} {Gpt-4 technical report}.
\newblock \emph{Preprint}, arXiv:2303.08774.

\bibitem[{Pfau et~al.(2024)Pfau, Merrill, and Bowman}]{pfau2024letsthinkdotdot}
Jacob Pfau, William Merrill, and Samuel~R. Bowman. 2024.
\newblock \href {https://arxiv.org/abs/2404.15758} {Let's think dot by dot: Hidden computation in transformer language models}.
\newblock \emph{Preprint}, arXiv:2404.15758.

\bibitem[{Qwen et~al.(2025)Qwen, :, Yang, Yang, Zhang, Hui, Zheng, Yu, Li, Liu, Huang, Wei, Lin, Yang, Tu, Zhang, Yang, Yang, Zhou, Lin, Dang, Lu, Bao, Yang, Yu, Li, Xue, Zhang, Zhu, Men, Lin, Li, Tang, Xia, Ren, Ren, Fan, Su, Zhang, Wan, Liu, Cui, Zhang, and Qiu}]{qwen2025qwen25technicalreport}
Qwen, :, An~Yang, Baosong Yang, Beichen Zhang, Binyuan Hui, Bo~Zheng, Bowen Yu, Chengyuan Li, Dayiheng Liu, Fei Huang, Haoran Wei, Huan Lin, Jian Yang, Jianhong Tu, Jianwei Zhang, Jianxin Yang, Jiaxi Yang, Jingren Zhou, and 25 others. 2025.
\newblock \href {https://arxiv.org/abs/2412.15115} {Qwen2.5 technical report}.
\newblock \emph{Preprint}, arXiv:2412.15115.

\bibitem[{Rein et~al.(2024)Rein, Hou, Stickland, Petty, Pang, Dirani, Michael, and Bowman}]{rein2024gpqa}
David Rein, Betty~Li Hou, Asa~Cooper Stickland, Jackson Petty, Richard~Yuanzhe Pang, Julien Dirani, Julian Michael, and Samuel~R. Bowman. 2024.
\newblock \href {https://openreview.net/forum?id=Ti67584b98} {{GPQA}: A graduate-level google-proof q\&a benchmark}.
\newblock In \emph{First Conference on Language Modeling}.

\bibitem[{Saunshi et~al.(2024)Saunshi, Karp, Krishnan, Miryoosefi, Reddi, and Kumar}]{saunshi2024inductivebiasstackingimproving}
Nikunj Saunshi, Stefani Karp, Shankar Krishnan, Sobhan Miryoosefi, Sashank~J. Reddi, and Sanjiv Kumar. 2024.
\newblock \href {https://arxiv.org/abs/2409.19044} {On the inductive bias of stacking towards improving reasoning}.
\newblock \emph{Preprint}, arXiv:2409.19044.

\bibitem[{Shalev et~al.(2024)Shalev, Feder, and Goldstein}]{shalev2024distributionalreasoningllmsparallel}
Yuval Shalev, Amir Feder, and Ariel Goldstein. 2024.
\newblock \href {https://arxiv.org/abs/2406.13858} {Distributional reasoning in llms: Parallel reasoning processes in multi-hop reasoning}.
\newblock \emph{Preprint}, arXiv:2406.13858.

\bibitem[{Shao et~al.(2024)Shao, Wang, Zhu, Xu, Song, Bi, Zhang, Zhang, Li, Wu, and Guo}]{shao2024deepseekmathpushinglimitsmathematical}
Zhihong Shao, Peiyi Wang, Qihao Zhu, Runxin Xu, Junxiao Song, Xiao Bi, Haowei Zhang, Mingchuan Zhang, Y.~K. Li, Y.~Wu, and Daya Guo. 2024.
\newblock \href {https://arxiv.org/abs/2402.03300} {Deepseekmath: Pushing the limits of mathematical reasoning in open language models}.
\newblock \emph{Preprint}, arXiv:2402.03300.

\bibitem[{Shi et~al.(2025)Shi, Asi, Li, Yuan, Pan, Lee, and Xiao}]{shi2025swireasoningswitchthinkinglatentexplicit}
Dachuan Shi, Abedelkadir Asi, Keying Li, Xiangchi Yuan, Leyan Pan, Wenke Lee, and Wen Xiao. 2025.
\newblock \href {https://arxiv.org/abs/2510.05069} {Swireasoning: Switch-thinking in latent and explicit for pareto-superior reasoning llms}.
\newblock \emph{Preprint}, arXiv:2510.05069.

\bibitem[{Shinn et~al.(2023)Shinn, Cassano, Berman, Gopinath, Narasimhan, and Yao}]{shinn2023reflexionlanguageagentsverbal}
Noah Shinn, Federico Cassano, Edward Berman, Ashwin Gopinath, Karthik Narasimhan, and Shunyu Yao. 2023.
\newblock \href {https://arxiv.org/abs/2303.11366} {Reflexion: Language agents with verbal reinforcement learning}.
\newblock \emph{Preprint}, arXiv:2303.11366.

\bibitem[{Singh et~al.(2025)Singh, Fry, Perelman, Tart, Ganesh, El-Kishky, McLaughlin, Low, Ostrow, Ananthram, Nathan, Luo, Helyar, Madry, Efremov, Spyra, Baker-Whitcomb, Beutel, Karpenko, Makelov, Neitz, Wei, Barr, Kirchmeyer, Ivanov, Christakis, Gillespie, Tam, Bennett, Wan, Huang, Sandjideh, Yang, Kumar, Saraiva, Vallone, Gheorghe, Garcia, Braunstein, Liu, Schmidt, Mereskin, Mishchenko, Applebaum, Rogerson, Rajan, Wei, Kotha, Srivastava, Agrawal, Vijayvergiya, Tyra, Nair, Nayak, Eggers, Ji, Hoover, Chen, Chen, Barak, Minaiev, Hao, Baker, Lightcap, McKinzie, Wang, Quinn, Fioca, Hsu, Yang, Yu, Zhang, Brenner, Zetino, Raymond, Lugaresi, Paz, Hudson, Whitney, Li, Chen, Cole, Voss, Ding, Shen, Huang, Colby, Hallacy, Koch, Lu, Kaplan, Kim, Minott-Henriques, Frey, Yu, Czarnecki, Reid, Wei, Decareaux, Scheau, Zhang, Forbes, Tang, Goldberg, Roberts, Palmie, Kappler, Levine, Wright, Leo, Lin, Robinson, Grabb, Chen, Lim, Salama, Bhattacharjee, Tsipras, Li, Yu, Strouse, Williams, Hunn, Bayes, Arbus, Akyurek, Le,
  Widmann, Yani, Proehl, Sert, Cheung, Schwartz, Han, Jiang, Mitchell, Sigler, Wallace, Ritter, Kavanaugh, Mays, Nikishin, Li, Such, de~Avila Belbute~Peres, Raso, Bekerman, Tsimpourlas, Chantzis, Song, Zhang, Raila, McGrath, Briggs, Yang, Parascandolo, Chabot, Kim, Zhao, Valiant, Leclerc, Salman, Wang, Sheng, Jiang, Wang, Jin, Sikchi, Schmidt, Aspegren, Chen, Qiu, Lightman, Covert, Kivlichan, Silber, Sohl, Hammoud, Clavera, Lan, Akkaya, Kostrikov, Kofman, Etinger, Singal, Hehir, Huh, Pan, Wilczynski, Pachocki, Lee, Quinn, Kiros, Kalra, Samaroo, Wang, Wolfe, Chen, Wang, Harb, Han, Wang, Zhao, Chen, Yang, Tworek, Chand, Landon, Liang, Lin, Liu, Wang, Tang, Yin, Jang, Morris, Flynn, Ferstad, Heidecke, Fishbein, Hallman, Grant, Chien, Gordon, Park, Liss, Kraaijeveld, Guay, Mo, Lawson, McGrath, Vendrow, Jiao, Lee, Steele, Wang, Mao, Chen, Hayashi, Xiao, Salahi, Wu, Sekhri, Sharma, Singhal, Li, Nguyen, Gu-Lemberg, King, Liu, Stone, Yu, Ying, Georgiev, Lim, Tirumala, Miller, Ahmad, Lv, Clare, Fauconnet, Itow, Yang,
  Romaniuk, Anise, Byron, Pathak, Maksin, Lo, Ho, Jing, Wu, Xiong, Mamitsuka, Yang, McCallum, Held, Bourgeois, Engstrom, Kuhn, Feuvrier, Zhang, Switzer, Kondraciuk, Kaiser, Joglekar, Singh, Shah, Stratta, Williams, Chen, Sun, Cayton, Li, Zhang, Aljubeh, Nichols, Haines, Schwarzer, Gupta, Shah, Huang, Dong, Wang, Glaese, Carroll, Lampe, Malek, Sharman, Zhang, Wang, Pokrass, Florian, Pavlov, Wang, Chen, Wang, Feng, Bavarian, Lin, Abdool, Rohaninejad, Soto, Staudacher, LaFontaine, Marwell, Liu, Preston, Turley, Ansman, Blades, Pancha, Mikhaylin, Felix, Handa, Rai, Keskar, Brown, Nachum, Boiko, Murk, Watkins, Gleeson, Mishkin, Lesiewicz, Baltescu, Belov, Zhokhov, Pronin, Guo, Thacker, Liu, Yuan, Liu, Dias, Puckett, Arora, Mullapudi, Gaon, Miyara, Song, Aggarwal, Marsan, Yemiru, Xiong, Kshirsagar, Nuttall, Tsiupa, Eldan, Wang, James, Ziv, Shu, Nigmatullin, Jain, Talaie, Altman, Arnesen, Toizer, Toyer, Miserendino, Agarwal, Yoo, Heon, Ethersmith, Grove, Taylor, Bubeck, Banesiu, Amdo, Zhao, Wu, Santurkar, Zhao,
  Chaudhuri, Krishnaswamy, Shuaiqi, Xia, Cheng, Anadkat, Fishman, Tobin, Fu, Jain, Mei, Egoian, Kim, Golden, Mah, Lin, Imm, Sharpe, Yadlowsky, Choudhry, Eum, Sanjeev, Khan, Stramer, Wang, Xin, Gogineni, Christianson, Sanders, Patwardhan, Degry, Shadwell, Fu, Gao, Garipov, Sriskandarajah, Sherbakov, Kaftan, Hiratsuka, Wang, Song, Zhao, Peterson, Kharitonov, Chernova, Kosaraju, Kuo, Pong, Verma, Petrov, Jiang, Zhang, Zhou, Xie, Zhan, McCabe, DePue, Ellsworth, Bain, Thompson, Chen, Qi, Xiang, Shi, Dubois, Yu, Khakbaz, Wu, Qian, Lee, Chen, Zhang, Xiong, Tian, Cha, Bai, Yang, Yuan, Li, Zhang, Yang, Jin, Jiang, Wang, Wang, Liu, Stubenvoll, Dou, Wu, and Wang}]{singh2025openaigpt5card}
Aaditya Singh, Adam Fry, Adam Perelman, Adam Tart, Adi Ganesh, Ahmed El-Kishky, Aidan McLaughlin, Aiden Low, AJ~Ostrow, Akhila Ananthram, Akshay Nathan, Alan Luo, Alec Helyar, Aleksander Madry, Aleksandr Efremov, Aleksandra Spyra, Alex Baker-Whitcomb, Alex Beutel, Alex Karpenko, and 465 others. 2025.
\newblock \href {https://arxiv.org/abs/2601.03267} {Openai gpt-5 system card}.
\newblock \emph{Preprint}, arXiv:2601.03267.

\bibitem[{Su et~al.(2025)Su, Zhu, Xu, Jiao, Tian, and Zheng}]{su2025tokenassortedmixinglatent}
DiJia Su, Hanlin Zhu, Yingchen Xu, Jiantao Jiao, Yuandong Tian, and Qinqing Zheng. 2025.
\newblock \href {https://arxiv.org/abs/2502.03275} {Token assorted: Mixing latent and text tokens for improved language model reasoning}.
\newblock \emph{Preprint}, arXiv:2502.03275.

\bibitem[{Tan et~al.(2025)Tan, Li, Ju, Luo, Luan, and Song}]{tan2025thinksilentlythinkfast}
Wenhui Tan, Jiaze Li, Jianzhong Ju, Zhenbo Luo, Jian Luan, and Ruihua Song. 2025.
\newblock \href {https://arxiv.org/abs/2505.16552} {Think silently, think fast: Dynamic latent compression of llm reasoning chains}.
\newblock \emph{Preprint}, arXiv:2505.16552.

\bibitem[{Team et~al.(2025)Team, Du, Gao, Xing, Jiang, Chen, Li, Xiao, Du, Liao, Tang, Wang, Zhang, Yuan, Lu, Tang, Sung, Wei, Lai, Guo, Zhu, Ding, Hu, Yang, Zhang, Yao, Zhao, Lu, Li, Yu, Gao, Zheng, Yuan, Chen, Guo, Su, Wang, Zhao, Zhang, Liu, Yan, Wu, Shi, Ye, Yu, Dong, Zhang, Ma, Pan, Gong, Liu, Ma, Wei, Cao, Huang, Jiang, Gao, Xiong, He, Huang, Xu, Wu, He, Wei, Jia, Wu, Xu, Zu, Zhou, Pan, Charles, Li, Hu, Liu, Chen, Wang, Liu, Qin, Liu, Yang, Bao, Du, Wu, Wang, Zhou, Wang, Li, Zhu, Zhang, Wang, Yang, Huang, Huang, Xu, Yang, and Lin}]{kimiteam2025kimik15scalingreinforcement}
Kimi Team, Angang Du, Bofei Gao, Bowei Xing, Changjiu Jiang, Cheng Chen, Cheng Li, Chenjun Xiao, Chenzhuang Du, Chonghua Liao, Chuning Tang, Congcong Wang, Dehao Zhang, Enming Yuan, Enzhe Lu, Fengxiang Tang, Flood Sung, Guangda Wei, Guokun Lai, and 77 others. 2025.
\newblock \href {https://arxiv.org/abs/2501.12599} {Kimi k1.5: Scaling reinforcement learning with llms}.
\newblock \emph{Preprint}, arXiv:2501.12599.

\bibitem[{Touvron et~al.(2023)Touvron, Lavril, Izacard, Martinet, Lachaux, Lacroix, Rozière, Goyal, Hambro, Azhar, Rodriguez, Joulin, Grave, and Lample}]{touvron2023llamaopenefficientfoundation}
Hugo Touvron, Thibaut Lavril, Gautier Izacard, Xavier Martinet, Marie-Anne Lachaux, Timothée Lacroix, Baptiste Rozière, Naman Goyal, Eric Hambro, Faisal Azhar, Aurelien Rodriguez, Armand Joulin, Edouard Grave, and Guillaume Lample. 2023.
\newblock \href {https://arxiv.org/abs/2302.13971} {Llama: Open and efficient foundation language models}.
\newblock \emph{Preprint}, arXiv:2302.13971.

\bibitem[{Wang et~al.(2024{\natexlab{a}})Wang, Li, Shao, Xu, Dai, Li, Chen, Wu, and Sui}]{wang2024mathshepherdverifyreinforcellms}
Peiyi Wang, Lei Li, Zhihong Shao, R.~X. Xu, Damai Dai, Yifei Li, Deli Chen, Y.~Wu, and Zhifang Sui. 2024{\natexlab{a}}.
\newblock \href {https://arxiv.org/abs/2312.08935} {Math-shepherd: Verify and reinforce llms step-by-step without human annotations}.
\newblock \emph{Preprint}, arXiv:2312.08935.

\bibitem[{Wang et~al.(2025)Wang, Wang, Zhu, and Liu}]{wang2025system15reasoningtraversallanguage}
Xiaoqiang Wang, Suyuchen Wang, Yun Zhu, and Bang Liu. 2025.
\newblock \href {https://arxiv.org/abs/2505.18962} {System-1.5 reasoning: Traversal in language and latent spaces with dynamic shortcuts}.
\newblock \emph{Preprint}, arXiv:2505.18962.

\bibitem[{Wang et~al.(2024{\natexlab{b}})Wang, Caccia, Ostapenko, Yuan, Wang, and Sordoni}]{wang2024guidinglanguagemodelreasoning}
Xinyi Wang, Lucas Caccia, Oleksiy Ostapenko, Xingdi Yuan, William~Yang Wang, and Alessandro Sordoni. 2024{\natexlab{b}}.
\newblock \href {https://arxiv.org/abs/2310.05707} {Guiding language model reasoning with planning tokens}.
\newblock \emph{Preprint}, arXiv:2310.05707.

\bibitem[{Wang et~al.(2023)Wang, Wei, Schuurmans, Le, Chi, Narang, Chowdhery, and Zhou}]{wang2023selfconsistencyimproveschainthought}
Xuezhi Wang, Jason Wei, Dale Schuurmans, Quoc Le, Ed~Chi, Sharan Narang, Aakanksha Chowdhery, and Denny Zhou. 2023.
\newblock \href {https://arxiv.org/abs/2203.11171} {Self-consistency improves chain of thought reasoning in language models}.
\newblock \emph{Preprint}, arXiv:2203.11171.

\bibitem[{Wei et~al.(2023)Wei, Wang, Schuurmans, Bosma, Ichter, Xia, Chi, Le, and Zhou}]{wei2023chainofthoughtpromptingelicitsreasoning}
Jason Wei, Xuezhi Wang, Dale Schuurmans, Maarten Bosma, Brian Ichter, Fei Xia, Ed~Chi, Quoc Le, and Denny Zhou. 2023.
\newblock \href {https://arxiv.org/abs/2201.11903} {Chain-of-thought prompting elicits reasoning in large language models}.
\newblock \emph{Preprint}, arXiv:2201.11903.

\bibitem[{Wei et~al.(2025)Wei, Liu, Wu, and Fang}]{wei2025surveyfeedbackbasedmultistepreasoning}
Ting-Ruen Wei, Haowei Liu, Xuyang Wu, and Yi~Fang. 2025.
\newblock \href {https://arxiv.org/abs/2502.14333} {A survey on feedback-based multi-step reasoning for large language models on mathematics}.
\newblock \emph{Preprint}, arXiv:2502.14333.

\bibitem[{Wolf et~al.(2020)Wolf, Debut, Sanh, Chaumond, Delangue, Moi, Cistac, Rault, Louf, Funtowicz, Davison, Shleifer, von Platen, Ma, Jernite, Plu, Xu, Scao, Gugger, Drame, Lhoest, and Rush}]{wolf2020huggingfacestransformersstateoftheartnatural}
Thomas Wolf, Lysandre Debut, Victor Sanh, Julien Chaumond, Clement Delangue, Anthony Moi, Pierric Cistac, Tim Rault, Rémi Louf, Morgan Funtowicz, Joe Davison, Sam Shleifer, Patrick von Platen, Clara Ma, Yacine Jernite, Julien Plu, Canwen Xu, Teven~Le Scao, Sylvain Gugger, and 3 others. 2020.
\newblock \href {https://arxiv.org/abs/1910.03771} {Huggingface's transformers: State-of-the-art natural language processing}.
\newblock \emph{Preprint}, arXiv:1910.03771.

\bibitem[{Wu et~al.(2026)Wu, Teng, and Tu}]{wu2026parallelcontinuouschainofthoughtjacobi}
Haoyi Wu, Zhihao Teng, and Kewei Tu. 2026.
\newblock \href {https://arxiv.org/abs/2506.18582} {Parallel continuous chain-of-thought with jacobi iteration}.
\newblock \emph{Preprint}, arXiv:2506.18582.

\bibitem[{Wu et~al.(2025)Wu, Lu, Ren, Hu, Wu, Dai, and Wu}]{wu2025llmssinglethreadedreasonersdemystifying}
Junhong Wu, Jinliang Lu, Zixuan Ren, Gangqiang Hu, Zhi Wu, Dai Dai, and Hua Wu. 2025.
\newblock \href {https://arxiv.org/abs/2508.03440} {Llms are single-threaded reasoners: Demystifying the working mechanism of soft thinking}.
\newblock \emph{Preprint}, arXiv:2508.03440.

\bibitem[{Xu et~al.(2025)Xu, Guo, Zeng, and Miao}]{xu2025softcotsoftchainofthoughtefficient}
Yige Xu, Xu~Guo, Zhiwei Zeng, and Chunyan Miao. 2025.
\newblock \href {https://arxiv.org/abs/2502.12134} {Softcot: Soft chain-of-thought for efficient reasoning with llms}.
\newblock \emph{Preprint}, arXiv:2502.12134.

\bibitem[{Yang et~al.(2025{\natexlab{a}})Yang, Li, Yang, Zhang, Hui, Zheng, Yu, Gao, Huang, Lv, Zheng, Liu, Zhou, Huang, Hu, Ge, Wei, Lin, Tang, Yang, Tu, Zhang, Yang, Yang, Zhou, Zhou, Lin, Dang, Bao, Yang, Yu, Deng, Li, Xue, Li, Zhang, Wang, Zhu, Men, Gao, Liu, Luo, Li, Tang, Yin, Ren, Wang, Zhang, Ren, Fan, Su, Zhang, Zhang, Wan, Liu, Wang, Cui, Zhang, Zhou, and Qiu}]{yang2025qwen3technicalreport}
An~Yang, Anfeng Li, Baosong Yang, Beichen Zhang, Binyuan Hui, Bo~Zheng, Bowen Yu, Chang Gao, Chengen Huang, Chenxu Lv, Chujie Zheng, Dayiheng Liu, Fan Zhou, Fei Huang, Feng Hu, Hao Ge, Haoran Wei, Huan Lin, Jialong Tang, and 41 others. 2025{\natexlab{a}}.
\newblock \href {https://arxiv.org/abs/2505.09388} {Qwen3 technical report}.
\newblock \emph{Preprint}, arXiv:2505.09388.

\bibitem[{Yang et~al.(2025{\natexlab{b}})Yang, Gribovskaya, Kassner, Geva, and Riedel}]{yang2025largelanguagemodelslatently}
Sohee Yang, Elena Gribovskaya, Nora Kassner, Mor Geva, and Sebastian Riedel. 2025{\natexlab{b}}.
\newblock \href {https://arxiv.org/abs/2402.16837} {Do large language models latently perform multi-hop reasoning?}
\newblock \emph{Preprint}, arXiv:2402.16837.

\bibitem[{Yao et~al.(2023)Yao, Zhao, Yu, Du, Shafran, Narasimhan, and Cao}]{yao2023reactsynergizingreasoningacting}
Shunyu Yao, Jeffrey Zhao, Dian Yu, Nan Du, Izhak Shafran, Karthik Narasimhan, and Yuan Cao. 2023.
\newblock \href {https://arxiv.org/abs/2210.03629} {React: Synergizing reasoning and acting in language models}.
\newblock \emph{Preprint}, arXiv:2210.03629.

\bibitem[{{Yentinglin}(2025)}]{yentinglin2025aime}
{Yentinglin}. 2025.
\newblock \href {https://huggingface.co/datasets/yentinglin/aime_2025} {{AIME 2025: American Invitational Mathematics Examination}}.
\newblock Hugging Face Dataset.

\bibitem[{Yu et~al.(2025)Yu, Jiang, Kang, Hao, and Qin}]{yu2025flowreasoningtrainingllms}
Fangxu Yu, Lai Jiang, Haoqiang Kang, Shibo Hao, and Lianhui Qin. 2025.
\newblock \href {https://arxiv.org/abs/2406.05673} {Flow of reasoning: Training llms for divergent reasoning with minimal examples}.
\newblock \emph{Preprint}, arXiv:2406.05673.

\bibitem[{Zhang et~al.(2025{\natexlab{a}})Zhang, Zhu, Sun, Luo, Qiao, Du, Zheng, Chen, and Zhang}]{zhang2025lightthinkerthinkingstepbystepcompression}
Jintian Zhang, Yuqi Zhu, Mengshu Sun, Yujie Luo, Shuofei Qiao, Lun Du, Da~Zheng, Huajun Chen, and Ningyu Zhang. 2025{\natexlab{a}}.
\newblock \href {https://arxiv.org/abs/2502.15589} {Lightthinker: Thinking step-by-step compression}.
\newblock \emph{Preprint}, arXiv:2502.15589.

\bibitem[{Zhang et~al.(2025{\natexlab{b}})Zhang, He, Yan, Shen, Zhao, Wang, Shen, and Wang}]{zhang2025softthinkingunlockingreasoning}
Zhen Zhang, Xuehai He, Weixiang Yan, Ao~Shen, Chenyang Zhao, Shuohang Wang, Yelong Shen, and Xin~Eric Wang. 2025{\natexlab{b}}.
\newblock \href {https://arxiv.org/abs/2505.15778} {Soft thinking: Unlocking the reasoning potential of llms in continuous concept space}.
\newblock \emph{Preprint}, arXiv:2505.15778.

\bibitem[{Zheng et~al.(2024)Zheng, Liu, Xie, Li, and Li}]{zheng2024progressivehintpromptingimprovesreasoning}
Chuanyang Zheng, Zhengying Liu, Enze Xie, Zhenguo Li, and Yu~Li. 2024.
\newblock \href {https://arxiv.org/abs/2304.09797} {Progressive-hint prompting improves reasoning in large language models}.
\newblock \emph{Preprint}, arXiv:2304.09797.

\bibitem[{Zhou et~al.(2023)Zhou, Schärli, Hou, Wei, Scales, Wang, Schuurmans, Cui, Bousquet, Le, and Chi}]{zhou2023leasttomostpromptingenablescomplex}
Denny Zhou, Nathanael Schärli, Le~Hou, Jason Wei, Nathan Scales, Xuezhi Wang, Dale Schuurmans, Claire Cui, Olivier Bousquet, Quoc Le, and Ed~Chi. 2023.
\newblock \href {https://arxiv.org/abs/2205.10625} {Least-to-most prompting enables complex reasoning in large language models}.
\newblock \emph{Preprint}, arXiv:2205.10625.

\end{thebibliography}
\begin{algorithm*}[h]
\caption{\textproc{SeLaR (Selective Latent Reasoning)}}
\label{alg:selar}
\begin{algorithmic}[1]
\Require Question $x_{1:n}$, model $\mathcal{M}$, max steps $S_{\max}$, top-$k$, entropy threshold $\tau$
\Ensure Answer $y_{1:m}$
\State \textbf{Init:} Embedding matrix $E$, max entropy $H_{\max} = \log k$
\For{$t=1$ to $S_{\max}$}
  \State $\ell_t \gets \mathcal{M}(x_{1:t-1})$;\quad $p_t \gets \mathrm{softmax}(\ell_t)$ \Comment{Forward pass}
  \State $\mathcal{V}_k \gets \mathrm{top\text{-}k}(p_t)$ \Comment{Select top-$k$ tokens}
  \State $\hat{p}_t[v] \gets p_t[v] / \sum_{v' \in \mathcal{V}_k} p_t[v']$ for $v \in \mathcal{V}_k$ \Comment{Normalize over top-$k$}
  \State $H_t \gets -\sum_{v \in \mathcal{V}_k} \hat{p}_t[v] \log \hat{p}_t[v]$ \Comment{Compute entropy}
  \State $\bar{H}_t \gets H_t / H_{\max}$ \Comment{Normalize entropy to $[0, 1]$}
  \State $x_t \gets \mathrm{Sample}(p_t)$ \Comment{Sample discrete token for readable output}
  \State
  \If{$\bar{H}_t < \tau$} \Comment{\textbf{Deterministic Step:} Low entropy}
     \State $e_t \gets E[x_t]$ \Comment{Use discrete embedding}
  \Else \Comment{\textbf{Exploratory Step:} High entropy}
     \State $e_t \gets \sum_{v \in \mathcal{V}_k} \hat{p}_t[v] \cdot E[v]$ \Comment{Soft embedding}
     \color{Darkblue}
     \State $v^* \gets \arg\max_{v \in \mathcal{V}_k} p_t[v]$ \Comment{Dominant token}
     \State $\Delta_t \gets e_t - E[v^*]$ \Comment{Direction from dominant}
     \State $\hat{\Delta}_t \gets \Delta_t / (\|\Delta_t\| + \epsilon)$ \Comment{Unit direction}
     \State $\tilde{e}_t \gets e_t +  \bar{H}_t \cdot \hat{\Delta}_t \cdot \|\Delta_t\|$ \Comment{Contrastive regularization}
     \color{black}
  \EndIf
  \State Feed $e_t$ as input embedding for next step
  \If{$x_t = \text{\textless EOS\textgreater}$} \State \textbf{break} \EndIf
\EndFor
\State Extract answer $y$ from $x_{n+1:t}$
\State \Return $y$
\end{algorithmic}
\end{algorithm*}
\section*{Appendix}
\appendix

\section{Supplementary Details}
\subsection{SeLaR Implementation}
\label{supp_selar_imp}
Alg~\ref{alg:selar} provides the detailed implementation of SeLaR. The core selective activation mechanism is shown in black: at each step, we compute the normalized entropy over top-$k$ tokens and compare it against threshold $\tau$ to determine whether to use discrete embeddings (deterministic steps) or soft embeddings (exploratory steps). The {\color{Darkblue}contrastive regularization} component is outlined in \textcolor{Darkblue}{blue}, which pushes the soft embedding away from the dominant token proportionally to the entropy, preventing premature collapse.
\begin{table*}[t]
\centering
\caption{Sensitivity analysis on Qwen3-8B. We vary the entropy threshold $\tau$ and top-$k$ value while keeping other settings fixed.}
\label{tab:sensitivity}
\begin{tabular}{cc|ccccc|c}
\toprule
$k$ & $\tau$ & GSM8K & MATH500 & GPQA & AIME24 & AIME25 & Avg \\
\midrule
\multicolumn{8}{c}{\cellcolor{lightblue}\textit{Varying $\tau$ (fixed $k=3$)}} \\
3 & 0.3 & 95.00 & 96.40 & 53.03 & 76.67 & 80.00 & 80.22 \\
3 & 0.4 & 95.22 & 96.60 & 54.55 & \textbf{83.33} & 70.00 & 79.94 \\
3 & 0.5 & 95.60 & \textbf{97.00} & 55.05 & 76.67 & \textbf{80.00} & \textbf{80.86} \\
3 & 0.6 & \textbf{95.83} & 96.00 & \textbf{60.10} & 76.67 & 70.00 & 79.72 \\
3 & 0.7 & 95.53 & 96.40 & \textbf{60.10} & 76.67 & 56.67 & 77.07 \\
\midrule
\multicolumn{8}{c}{\cellcolor{lightblue}\textit{Varying $k$ (fixed $\tau=0.5$)}} \\
3 & 0.5 & \textbf{95.60} & \textbf{97.00} & 55.05 & \textbf{76.67} & \textbf{80.00} & \textbf{80.86} \\
5 & 0.5 & 95.30 & 96.40 & \textbf{61.62} & \textbf{76.67} & 53.33 & 76.66 \\
7 & 0.5 & 95.00 & 96.60 & 55.56 & 73.33 & 63.33 & 76.76 \\
\bottomrule
\end{tabular}
\end{table*}
\subsection{Dataset Details}
\label{sec:datasets}

We evaluate our method on five reasoning benchmarks spanning mathematical problem-solving and knowledge-intensive question answering.

\paragraph{GSM8K}
is a benchmark for evaluating multi-step mathematical reasoning in natural language. Following standard practice, we evaluate on the official test set, which contains 1,319 grade-school level math word problems requiring explicit step-by-step reasoning. \hf: \url{https://huggingface.co/datasets/openai/gsm8k}.

\paragraph{MATH500}
is a challenging subset of the MATH dataset, consisting of 500 high school competition-level problems spanning algebra, geometry, number theory, and calculus. The problems require non-trivial symbolic manipulation and multi-step deductive reasoning. \hf: \url{https://huggingface.co/datasets/HuggingFaceH4/MATH-500}.

\paragraph{AIME 2024}
is a benchmark of 30 problems from the 2024 American Invitational Mathematics Examination. Each problem demands deep multi-step reasoning and precise numerical computation, with answers constrained to integers within a fixed range. \hf: \url{https://huggingface.co/datasets/HuggingFaceH4/aime_2024}.

\paragraph{AIME 2025}
is a benchmark of 30 problems from the 2025 AIME examination, featuring newly released competition problems with similar formats but increased novelty, providing a stringent test of generalization and reasoning robustness. \hf: \url{https://huggingface.co/datasets/yentinglin/aime_2025}.

\paragraph{GPQA Diamond}
is the most difficult split of the GPQA benchmark, containing 198 expert-curated questions across mathematics, physics, chemistry, biology, and computer science. The questions are designed to resist superficial pattern matching and require advanced domain knowledge and rigorous reasoning. \hf: \url{https://huggingface.co/datasets/hendrydong/gpqa_diamond_mc}.

\section{Sensitivity Analysis Details}
\label{app:Sensitivity Analysis Details}

Table~\ref{tab:sensitivity} presents the sensitivity analysis for SeLaR on Qwen3-8B.

\paragraph{Effect of Entropy Threshold $\tau$}
We vary $\tau$ from 0.3 to 0.7 with $k=3$. Lower thresholds activate latent reasoning too frequently, introducing perturbation at high-confidence steps, while higher thresholds activate it too conservatively, limiting exploration at true exploratory steps. The optimal $\tau$ varies across datasets (e.g., $\tau=0.4$–$0.7$), reflecting their inherent entropy characteristics: harder tasks benefit from reserving latent reasoning for highly uncertain steps, whereas tasks with more frequent exploratory steps favor earlier activation. Importantly, SeLaR remains stable across a wide range of thresholds ($\tau \in [0.3, 0.7]$), indicating that $\tau$ serves as a \textbf{coarse uncertainty gate} derived from the entropy distribution rather than a finely tuned hyperparameter.

\paragraph{Effect of Top-$k$ Value}
We vary $k$ from 3 to 7 while fixing $\tau=0.5$. Smaller $k$ values yield better average performance, with $k=3$ achieving 80.86\% compared to 76.66\% for $k=5$ and 76.76\% for $k=7$. This suggests that restricting soft embeddings to fewer high-probability candidates preserves semantic coherence, while larger $k$ values dilute the representation with low-probability tokens that introduce perturbation. 

\paragraph{Final Selection}
Based on the above analysis, we fix $k=3$ across all experiments and select dataset-specific thresholds that maximize individual benchmark performance: $\tau=0.6$ for GSM8K, $\tau=0.5$ for MATH500, $\tau=0.7$ for GPQA-Diamond, $\tau=0.4$ for AIME 2024, and $\tau=0.5$ for AIME 2025. These settings are used for all main results reported in Table~\ref{tab:bridge_results}.

\begin{figure*}[ht]
\centering
\includegraphics[width=\linewidth]{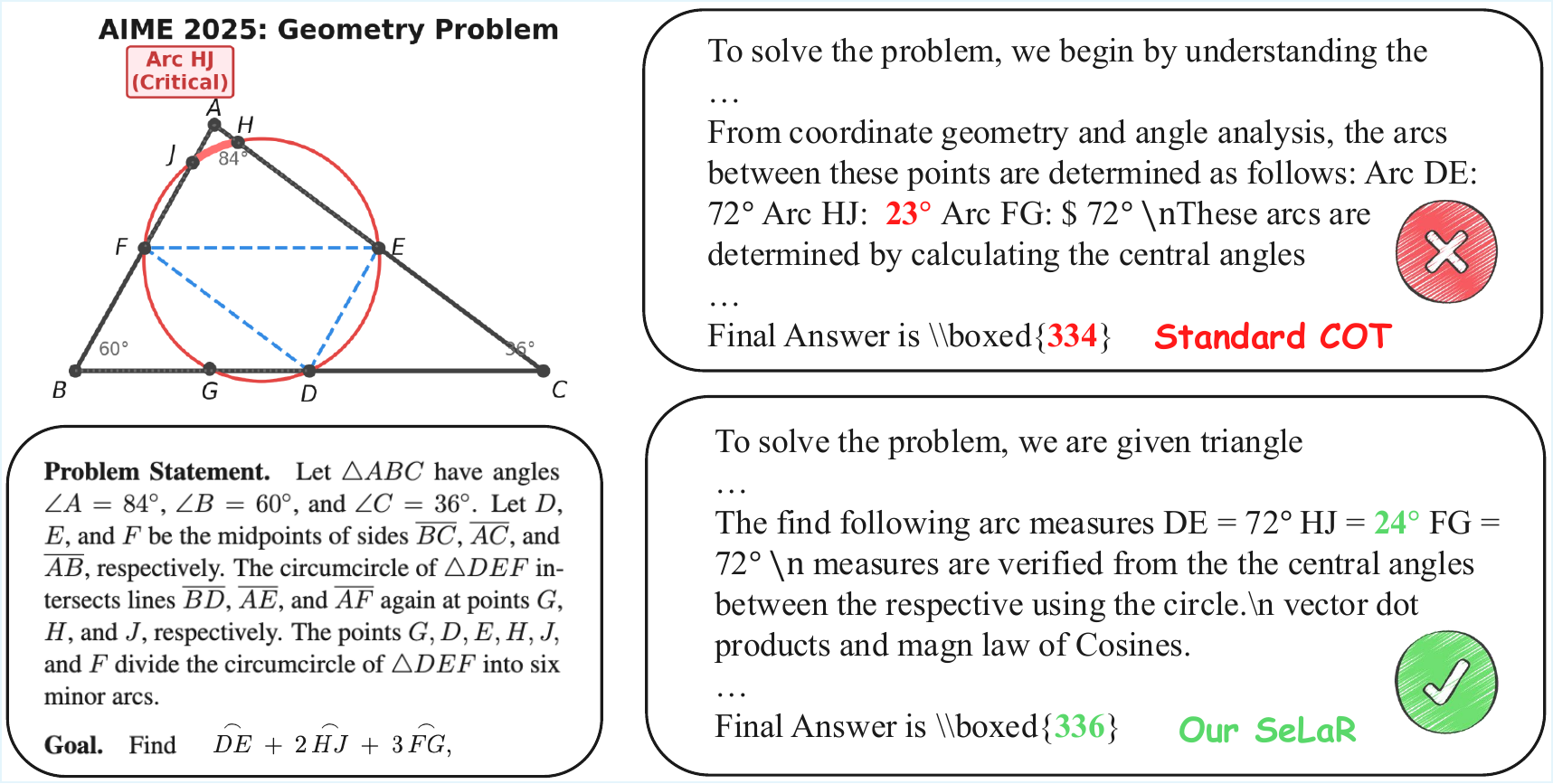}
\caption{Case study on an AIME 2025 geometry problem. Standard CoT computes \textcolor{red}{Arc HJ = 23°} at the critical exploratory step, leading to an incorrect final answer of \textcolor{red}{334}. SeLaR activates selective latent reasoning at this exploratory step, correctly computing \textcolor{Darkgreen}{Arc HJ = 24°} and yielding the correct answer \textcolor{Darkgreen}{336}.}
\label{fig:case_study}
\end{figure*}

\section{Experimental Results on Other Model Families}
\label{appdix:Deepseek}

\begin{figure}[ht]
\centering
\includegraphics[width=\columnwidth]{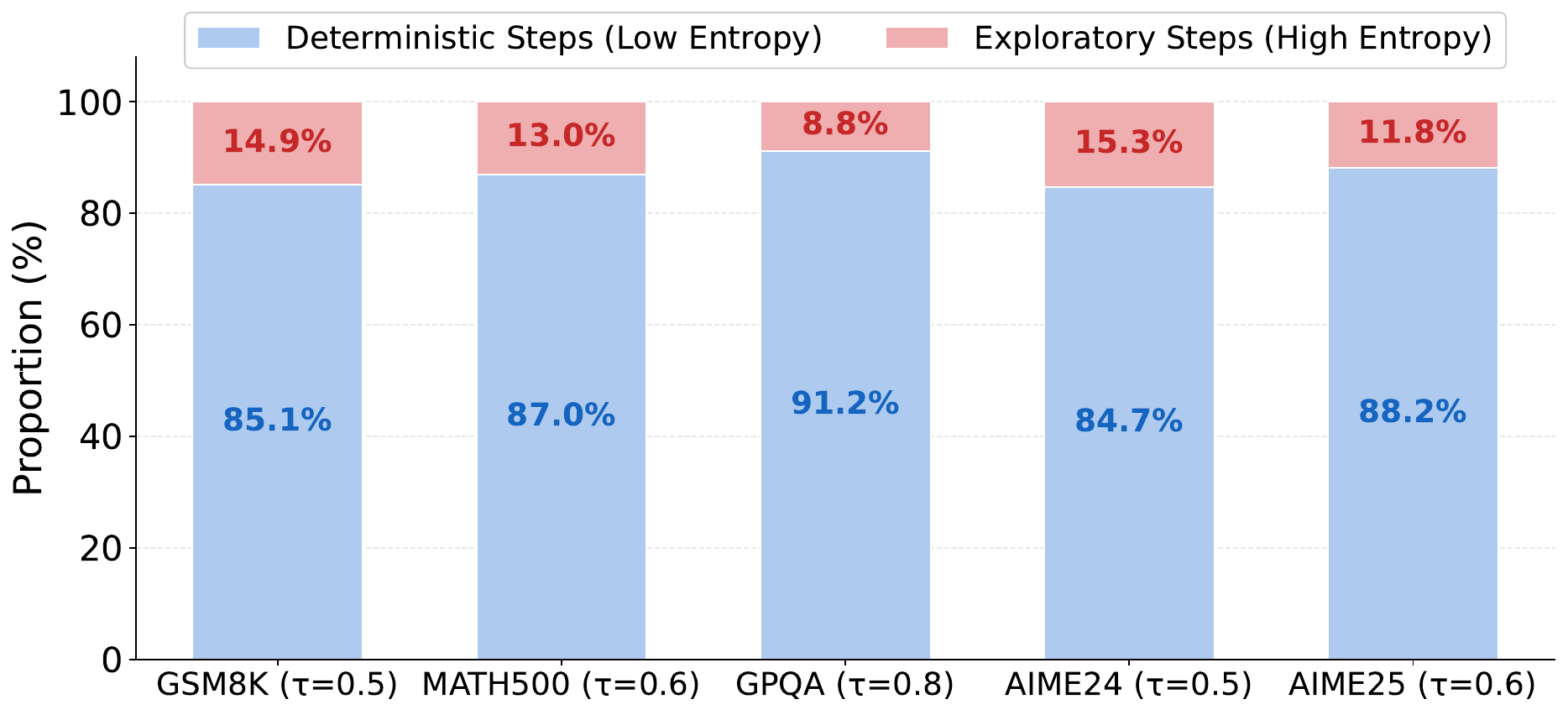}
\caption{Activation frequency analysis on DeepSeek-R1-Distill-Llama-8B. Exploratory steps account for 8.8\%--15.3\% of total reasoning tokens, higher than Qwen3-8B (6.2\%--13.8\%).}
\label{fig:activation_frequency_deepseek}
\end{figure}

\begin{table*}[t]
\centering
\caption{
Results of DeepSeek-R1-Distill-Llama-8B.
Results highlighted in \mycolorbox{lightgreen}{green} indicate performance comparable to or better than CoT (Sampling).
Results highlighted in \mycolorbox{lightred}{red} indicate a performance drop relative to CoT (Sampling).
}
\label{tab:deepseek_result}

\begin{tabular}{lcccccc}
\toprule
\multicolumn{1}{l|}{Method} &
GSM8K &
MATH500 &
GPQA &
AIME24 &
AIME25 &
Avg \\ \midrule

\multicolumn{1}{l|}{} &
\multicolumn{6}{c}{\cellcolor{lightblue}\textit{DeepSeek-R1-Distill-Llama-8B} \cite{deepseekai2025deepseekr1incentivizingreasoningcapability}} \\
\multicolumn{1}{l|}{CoT (Sampling)} & 
89.01 & 90.20 & 42.93 & 36.67 & 30.00 & {\color[HTML]{333333}57.76} \\
\multicolumn{1}{l|}{CoT (Greedy)} & 
\cellcolor{lightred}85.29 &
\cellcolor{lightred}83.60 &
\cellcolor{lightred}30.30 &
\cellcolor{lightred}30.00 &
\cellcolor{lightred}26.67 &
\cellcolor{lightred}51.17 \\
\multicolumn{1}{l|}{Soft Thinking} &
\cellcolor{lightred}85.67 &
\cellcolor{lightred}82.20 &
\cellcolor{lightred}32.83 &
\cellcolor{lightred} 30.00&
\cellcolor{lightred}23.33 &
\cellcolor{lightred}50.81 \\
\multicolumn{1}{l|}{SwiR} &
\cellcolor{lightgreen}89.31 &
\cellcolor{lightred}87.80 &
\cellcolor{lightgreen}\textbf{45.96} &
\cellcolor{lightgreen}\textbf{50.00} &
\cellcolor{lightred}23.33 &
\cellcolor{lightgreen}59.28 \\
\multicolumn{1}{l|}{\textbf{SeLaR}} &
\cellcolor{lightgreen}\textbf{90.22} &
\cellcolor{lightred}\textbf{88.20} &
\cellcolor{lightred} 40.91&
\cellcolor{lightgreen}46.67 &
\cellcolor{lightgreen}\textbf{36.67} &
\cellcolor{lightgreen}\textbf{60.53} \\
\bottomrule
\end{tabular}
\end{table*}

Table~\ref{tab:deepseek_result} presents results on DeepSeek-R1-Distill-Llama-8B, a model from a different family from Qwen3. SeLaR achieves the highest average accuracy (60.53\%), outperforming CoT (Sampling) by 2.77\% and SwiR by 1.25\%. However, the improvements are less pronounced compared to Qwen3 models. As shown in Figure~\ref{fig:activation_frequency_deepseek}, DeepSeek-R1-Distill-Llama-8B exhibits higher activation frequencies (8.8\%--15.3\%) compared to Qwen3-8B (6.2\%--13.8\%), indicating lower confidence during reasoning. This triggers more frequent exploratory steps and introduces additional perturbation, limiting the effectiveness of selective activation. This sensitivity is a natural trade-off of the training-free design: SeLaR operates at the embedding level without modifying hidden states and therefore inevitably depends on the base model’s intrinsic reasoning capability in hidden-state space.

\section{Case Study}
Figure~\ref{fig:case_study} presents a qualitative comparison on an AIME 2025 geometry problem. Both Standard CoT and SeLaR follow identical reasoning paths initially, but diverge at a critical exploratory point: computing Arc HJ. Standard CoT commits to 23° and arrives at an incorrect answer of 334, while SeLaR correctly computes 24° and yields the correct answer 336.

\end{document}